\documentclass[11pt,a4paper]{article}
\usepackage[hyperref]{emnlp-ijcnlp-2019}
\usepackage{times}
\usepackage{latexsym}
\usepackage{amsmath}
\usepackage{amssymb}
\usepackage{color,colortbl}
\usepackage{booktabs}
\usepackage{subcaption}
\usepackage{multirow}
\usepackage{ifthen}
\usepackage{xspace}
\usepackage{graphicx}
\usepackage{framed}
\usepackage{textcomp}
\usepackage{float}
\usepackage{makecell}
\usepackage{booktabs}
\usepackage{comment}
\usepackage{url}
\usepackage{ulem}
\usepackage{stackengine}
\usepackage{enumerate}
\aclfinalcopy % Uncomment this line for the final submission

\newcommand\sC{\ensuremath{\mathcal{C}}}

\newcommand\sF{\ensuremath{\mathcal{F}}}
\newcommand\sG{\ensuremath{\mathcal{G}}}

\newcommand\sO{\ensuremath{\mathcal{O}}}

\newcommand\sX{\ensuremath{\mathcal{X}}}
\newcommand\sY{\ensuremath{\mathcal{Y}}}

\newcommand\R{\ensuremath{\mathbb{R}}} % Real numbers
\newcommand{\1}{\mathbb{I}} % Indicator (don't use \mathbbm{1} because bbm is not TrueType)
\newcommand\refeqn[1]{(\ref{eqn:#1})}

\newcommand\refsec[1]{Section~\ref{sec:#1}}

\newcommand\reffig[1]{Figure~\ref{fig:#1}}

\newcommand\reftab[1]{Table~\ref{tab:#1}}
\newcommand\refapp[1]{Appendix~\ref{sec:#1}}

\ifthenelse{\isundefined{\definition}}{\newtheorem{definition}{Definition}}{}
\ifthenelse{\isundefined{\assumption}}{\newtheorem{assumption}{Assumption}}{}
\ifthenelse{\isundefined{\hypothesis}}{\newtheorem{hypothesis}{Hypothesis}}{}
\ifthenelse{\isundefined{\proposition}}{\newtheorem{proposition}{Proposition}}{}
\ifthenelse{\isundefined{\theorem}}{\newtheorem{theorem}{Theorem}}{}
\ifthenelse{\isundefined{\lemma}}{\newtheorem{lemma}{Lemma}}{}
\ifthenelse{\isundefined{\corollary}}{\newtheorem{corollary}{Corollary}}{}
\ifthenelse{\isundefined{\alg}}{\newtheorem{alg}{Algorithm}}{}
\ifthenelse{\isundefined{\example}}{\newtheorem{example}{Example}}{}
\newcommand\nl[1]{``\textit{#1}''}
\newcommand\ball[1]{\ensuremath{B_\text{perturb}({#1})}}

\newcommand\finput{g^{\text{word}}}
\newcommand\relu{\operatorname{ReLU}}
\newcommand\vpre{\phi^{\text{pre}}}

\newcommand \zin{\ensuremath{z^\text{dep}}}
\newcommand \zout{\ensuremath{z^\text{res}}}
\newcommand \oin{\ensuremath{\sO^\text{dep}}}
\newcommand \oout{\ensuremath{\sO^\text{res}}}
\newcommand \lin{\ensuremath{\ell^\text{dep}}}
\newcommand \lout{\ensuremath{\ell^\text{res}}}
\newcommand \uin{\ensuremath{u^\text{dep}}}
\newcommand \uout{\ensuremath{u^\text{res}}}
\newcommand \dep{\operatorname{dep}}

\newcommand{\bowimdb}{\textsc{BoW}\xspace}
\newcommand{\bowsnli}{\textsc{BoW}\xspace}
\newcommand{\decompattn}{\textsc{DecompAttn}\xspace}

\title{Certified Robustness to Adversarial Word Substitutions}

\author{
  Robin Jia \qquad Aditi Raghunathan \qquad Kerem G\"oksel \qquad Percy Liang \\
  Computer Science Department, Stanford University \\
  \texttt{\{robinjia,aditir,kerem,pliang\}@cs.stanford.edu}
}

\date{}

\begin{document}

\maketitle
\begin{abstract}
  State-of-the-art NLP models can often be fooled by adversaries that apply
seemingly innocuous label-preserving transformations (e.g., paraphrasing) to input text.
The number of possible transformations scales exponentially with text length,
so data augmentation cannot cover all transformations of an input.
This paper considers one exponentially large family of label-preserving transformations,
in which every word in the input can be replaced with a similar word.
We train the first models that are provably robust to \emph{all} word substitutions in this family.
Our training procedure uses Interval Bound Propagation (IBP) to minimize an upper bound on the worst-case loss that any combination of word substitutions can induce.
To evaluate models' robustness to these transformations, we measure accuracy on adversarially chosen word substitutions applied to test examples.
Our IBP-trained models attain $75\%$ adversarial accuracy on both 
sentiment analysis on IMDB and natural language inference on SNLI.
In comparison, on IMDB, models trained normally and ones trained with data augmentation
achieve adversarial accuracy of only $8\%$ and $35\%$, respectively. 

\end{abstract}

\section{Introduction}
Machine learning models have achieved impressive accuracy on many
NLP tasks, but they are surprisingly brittle.
Adding distracting text to the input
\cite{jia2017adversarial}, 
paraphrasing the text \cite{iyyer2018adversarial,ribeiro2018sears},
replacing words with similar words \cite{alzantot2018adversarial},
or inserting character-level ``typos'' \cite{belinkov2017synthetic, ebrahimi2017hotflip} can
significantly degrade a model's performance. 
Such perturbed inputs are called \emph{adversarial
  examples}, and have shown to break models in other domains as well, most notably in vision \cite{szegedy2014intriguing,
goodfellow2015explaining}.
Since humans are not fooled by the same perturbations,
the widespread existence of adversarial examples exposes troubling gaps in models' understanding.

\begin{figure}
\center
\vspace{-15px}
\includegraphics[width=\linewidth]{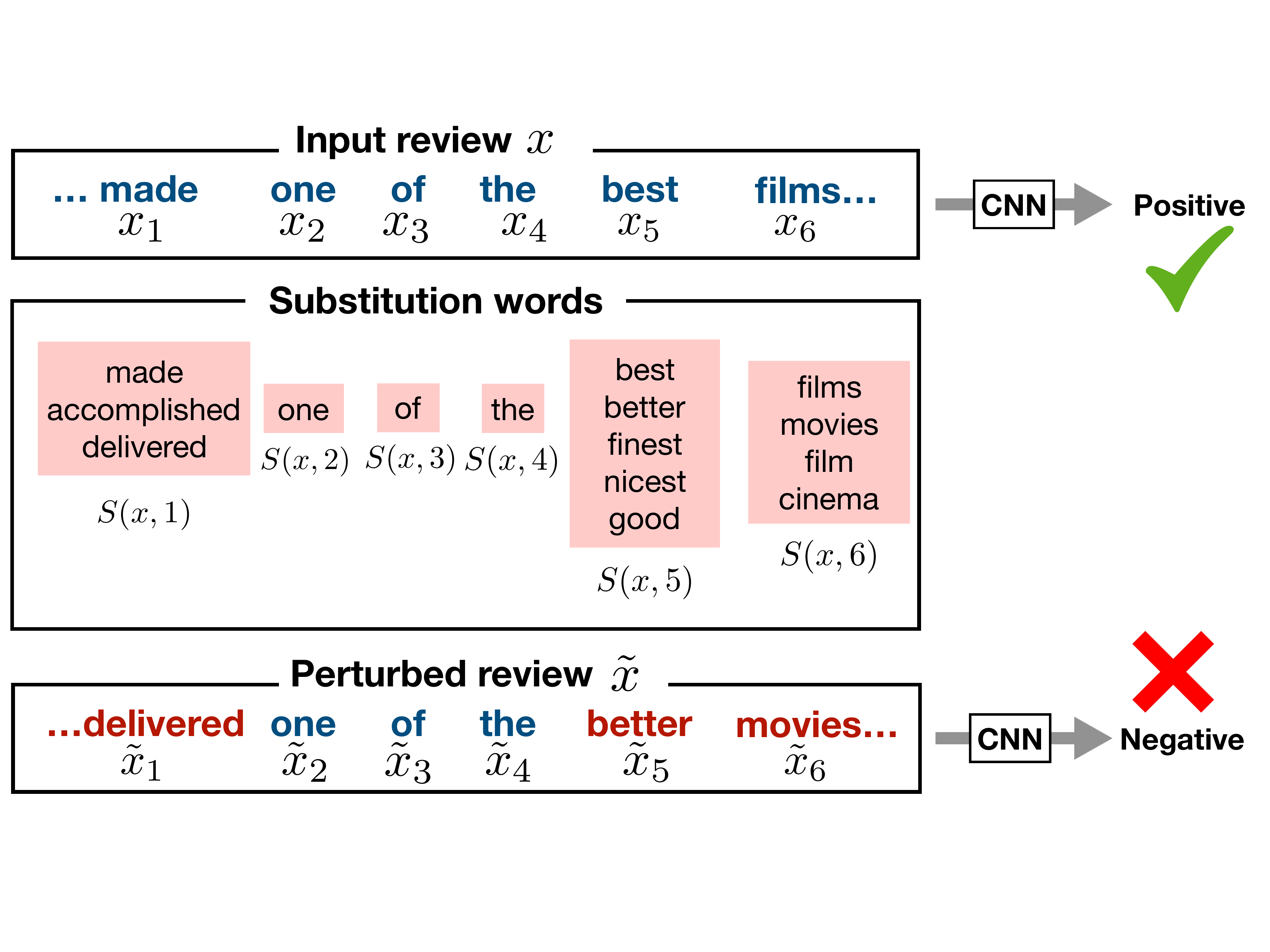}
\vspace{-30px}
  \caption{
    Word substitution-based perturbations in sentiment analysis. 
    For an input $x$, we consider perturbations $\tilde{x}$,
    in which \emph{every} word $x_i$ can be replaced with any similar word from the set $S(x, i)$, 
    without changing the original sentiment. 
    Models can be easily fooled by adversarially chosen perturbations
    (e.g., changing \nl{best} to \nl{better}, \nl{made} to \nl{delivered},
    \nl{films} to \nl{movies}),
    but the ideal model would be robust to all combinations of word substitutions.
  }
\label{fig:attack}
\end{figure}

In this paper, we focus on the word substitution perturbations of \citet{alzantot2018adversarial}.
In this setting, an attacker may replace every word in the input with a similar word (that ought not to change the label), leading to an exponentially large number of possible perturbations.
\reffig{attack} shows an example of these word substitutions.
As demonstrated by a long line of work in computer vision,
it is challenging to make models that are robust
to very large perturbation spaces,
even when the set of perturbations is known at training time
\cite{goodfellow2015explaining,athalye2018obfuscated,raghunathan2018certified,wong2018provable}.

Our paper addresses two key questions.
First, is it possible to guarantee that a model is robust against \emph{all} adversarial perturbations of a given input?
Existing methods that use heuristic search to attack models
\cite{ebrahimi2017hotflip,alzantot2018adversarial}
are slow and cannot provide guarantees of robustness,
since the space of possible perturbations is too large to search exhaustively.
We obtain guarantees by leveraging Interval Bound Propagation (IBP),
a technique that was previously applied to feedforward networks and CNNs in computer vision \cite{dvijotham2018training}.
IBP efficiently computes a tractable \emph{upper bound} on the loss of the worst-case perturbation.
When this upper bound on the worst-case loss is small, the model is guaranteed to be robust 
to all perturbations, providing a \emph{certificate} of robustness. 
To apply IBP to NLP settings,
we derive new interval bound formulas for multiplication and softmax layers,
which enable us to compute IBP bounds for LSTMs \cite{hochreiter1997lstm} and attention layers \cite{bahdanau2015neural}.
We also extend IBP to handle discrete perturbation sets, rather than the continuous ones used in vision.

Second, can we train models that are robust in this way?
Data augmentation can sometimes mitigate the effect of adversarial examples
\cite{jia2017adversarial,belinkov2017synthetic,ribeiro2018sears,liu2019inoculation}, 
but it is insufficient when considering very large perturbation spaces \cite{alzantot2018adversarial}.
Adversarial training strategies from computer vision \citep{madry2018towards} rely on gradient information,
and therefore do not extend to the discrete perturbations seen in NLP.
We instead use \emph{certifiably robust training}, in which we train models to optimize the IBP upper bound \cite{dvijotham2018training}.

We evaluate certifiably robust training on two tasks---sentiment analysis on the IMDB dataset \citep{maas2011imdb}
and natural language inference on the SNLI dataset \citep{bowman2015large}.
Across various model architectures (bag-of-words, CNN, LSTM, and attention-based),
certifiably robust training consistently yields models which are provably robust to all perturbations
on a large fraction of test examples. 
A normally-trained model has only $8\%$ and $41\%$ accuracy on IMDB and SNLI,
respectively, when evaluated on adversarially perturbed test examples.
With certifiably robust training, we achieve $75\%$ adversarial accuracy for both IMDB and SNLI.
Data augmentation fares much worse than certifiably robust training,
with adversarial accuracies falling to $35\%$ and $71\%$, respectively.

\section{Setup}
\label{sec:setup}
We consider tasks where a model must predict a label $y \in \sY$ given textual input $x \in \sX$.
For example, for sentiment analysis,
the input $x$ is a sequence of words $x_1, x_2, \dots, x_L$, and the goal is to assign a
label $y \in \{-1, 1\}$ denoting negative or positive sentiment, respectively.
We use $z = (x, y)$ to denote an example with input $x$ and label $y$,
and use $\theta$ to denote parameters of a model.
Let $f(z, \theta) \in \R$ denote some loss of a model with parameters $\theta$ on
example $z$. We evaluate models on $f^\text{0-1}(z, \theta)$, the zero-one loss under model $\theta$.

\subsection{Perturbations by word substitutions}
\label{sec:perturb}
Our goal is to build models that are robust to label-preserving perturbations.
In this work, we focus on perturbations where words of the input are substituted with similar words.
Formally, for every word $x_i$, we consider a set of allowed substitution words $S(x, i)$,
including $x_i$ itself.
We use $\tilde{x}$ to denote a perturbed version of $x$, 
where each word $\tilde{x}_i$ is in $S(x, i)$.
For an example $z = (x, y)$, let $\ball{z}$ denote the set of \emph{all} allowed perturbations of $z$:
\begin{align}
\label{eqn:ball}
\ball{z} &= \{ (\tilde{x}, y): \tilde{x}_i \in S(x, i) ~~ \forall i \}.
\end{align}
\reffig{attack} provides an illustration of word substitution perturbations. 
We choose $S(x, i)$ so that $\tilde{x}$
is likely to be grammatical and have the same label as $x$ (see \refsec{exp-setup}).

\subsection{Robustness to all perturbations}
\label{sec:cert}
Let $\sF(z, \theta)$ denote the set of losses of the network on the set of perturbed examples
defined in \refeqn{ball}: 
\begin{align}
\sF(z, \theta) &= \{ f(\tilde{z}, \theta): \tilde{z} \in \ball{z} \}.
\end{align}
We define the \emph{robust loss} as $\max \sF(z, \theta)$, the loss due to worst-case perturbation.
A model is robust at $z$ if it classifies all inputs in the perturbation set correctly, 
i.e., the robust zero-one loss $\max \sF^\text{0-1}(z, \theta) = 0$.
Unfortunately, the robust loss is often intractable to compute,
as each word can be perturbed independently.
For example, reviews in the IMDB dataset \citep{maas2011imdb} 
have a median of $10^{31}$ possible perturbations and max of $10^{271}$, 
far too many to enumerate.
We instead propose a tractable \emph{upper bound} by
constructing a set $\sO(z, \theta) \supseteq \sF(z, \theta)$.
Note that 
\begin{align}
\max \sO^\text{0-1}(z, \theta) = 0 &\Rightarrow \max \sF^\text{0-1}(z, \theta) = 0 \nonumber \\
                       &\Leftrightarrow \text{robust at $z$}.
\end{align}
Therefore, whenever $\max \sO^\text{0-1}(z, \theta) = 0$,
this fact is sufficient to \emph{certify} robustness
to all perturbed examples $\ball{z}$. However, since $\sO^\text{0-1}(z, \theta) \supseteq \sF^\text{0-1}(z, \theta)$,
the model could be robust even if $\max \sO^{\text{0-1}}(z, \theta) \ne 0$.

\section{Certification via Interval Bound Propagation}
\label{sec:ibp}
We now show how to use Interval Bound Propagation (IBP) \cite{dvijotham2018training} to obtain a 
superset $\sO(z, \theta)$ of the losses of perturbed inputs $\sF(z, \theta)$,
given $z$, $\theta$, and $\ball{z}$.
For notational convenience, we drop $z$ and $\theta$.
The key idea is to compute upper and lower bounds on the activations in each layer of the network, in terms of bounds computed for previous layers.
These bounds \emph{propagate} through the network, as in a standard forward pass,
until we obtain bounds on the final output, i.e., the loss $f$.
While IBP bounds may be loose in general,
\refsec{results} shows that training networks to minimize the upper bound on $f$
makes these bounds much tighter
\cite{gowal2018effectiveness,raghunathan2018certified}. 

Formally, let $g^i$ denote a scalar-valued function of $z$ and $\theta$ (e.g., a single activation in one layer of the network) computed at node $i$ of the computation graph for a given network.
Let $\dep(i)$ be the set of nodes used to compute $g^i$ in the computation graph (e.g., activations of the previous layer).
Let $\sG^i$ denote the set of possible values
of $g^i$ across all examples in $\ball{z}$.
We construct an interval $\sO^i = [\ell^i, u^i]$ 
that contains all these possible values of $g^i$, i.e., $\sO^i \supseteq \sG^i$.
$\sO^i$ is computed from the intervals $\sO^{\dep(i)}= \{\sO^j: j \in \dep(i) \}$ 
of the dependencies of $g^i$.
Once computed, $\sO^i$ can then be used to compute intervals on nodes that depend on $i$.
In this way, bounds propagate through the entire computation graph in an efficient forward pass.

We now discuss how to compute interval bounds for NLP models and word substitution perturbations.
We obtain interval bounds for model inputs 
given $\ball{z}$ (\refsec{input}), then show
how to compute $\sO^i$ from $\sO^\text{dep(i)}$ for 
elementary operations used in standard NLP models (\refsec{elem}).
Finally, we use these bounds to certify robustness
and train robust models.

\subsection{Bounds for the input layer}
\label{sec:input}

\begin{figure}
\center
\includegraphics[width=\linewidth]{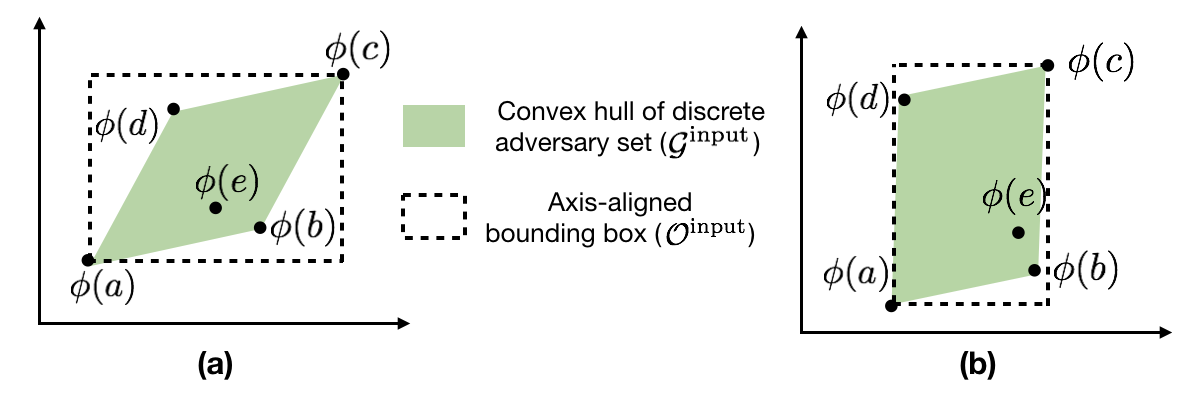}
  \caption{Bounds on the word vector inputs to the neural network. Consider a word (sentence of length one) $x = a$ with the set of substitution words $S(x, 1) = \{a, b, c, d, e\}$. 
  (a) IBP constructs axis-aligned bounds around a set of word vectors. These bounds may be loose, especially if the word vectors are pre-trained and fixed.
  (b) A different word vector space can give tighter IBP bounds, if the convex hull of the word vectors is better approximated by an axis-aligned box.
  }
\label{fig:transform}
\end{figure}
Previous work~\cite{gowal2018effectiveness} applied IBP to continuous image perturbations, which are naturally represented with interval bounds \cite{dvijotham2018training}.
We instead work with discrete word substitutions, which we must convert into interval bounds $\sO^\text{input}$ in order to use IBP.
Given input words $x = x_1, \dots, x_L$, 
we assume that the model embeds each word as $g^\text{input} = [\phi(x_1), \dotsc, \phi(x_L)] \in \R^{L \times d}$,
where $\phi(x_i) \in \mathbb{R}^d$ is the word vector for word $x_i$.
To compute $\sO^\text{input} \supseteq \sG^\text{input}$,
recall that each input word $x_i$ can be replaced with any $\tilde{x}_i \in S(x, i)$.
So, 
for each coordinate $j \in \{1, \dotsc, d\}$,
we can obtain an interval bound $\sO^\text{input}_{ij} = [\ell^\text{input}_{ij}, u^\text{input}_{ij}]$ 
for $g^\text{input}_{ij}$
by computing the smallest axis-aligned box that contains all the word vectors:
\begin{align}
  \label{eqn:input}
  \ell^\text{input}_{ij} &= \min \limits_{w \in S(x, i)} \phi(w)_j,  ~~ 
  u^\text{input}_{ij} &= \max \limits_{w \in S(x, i)} \phi(w)_j. 
\end{align}
\reffig{transform} illustrates these bounds.
We can view this as relaxing a set of discrete points to a convex set that contains all of the points.
\refsec{models} discusses modeling choices to make this box tighter.

\subsection{Interval bounds for elementary functions}
\label{sec:elem}
Next, we describe how to compute the interval of a node $i$ from intervals of its dependencies. 
\citet{gowal2018effectiveness} show how to efficiently compute interval bounds for affine transformations (i.e., linear layers) and monotonic elementwise nonlinearities
(see \refapp{ibp}). 
This suffices to compute interval bounds for feedforward networks and CNNs.
However, common NLP model components like LSTMs and attention also rely on softmax (for attention),
element-wise multiplication (for LSTM gates), and dot product 
(for computing attention scores).
We show how to compute interval bounds for these new operations.
These building blocks can be used to compute interval bounds not only for LSTMs and attention,
but also for any model that uses these elementary functions.

For ease of notation, we drop the superscript $i$ on $g^i$ and write that a node computes a result 
$\zout = g(\zin)$
where $\zout \in \R$ and $\zin \in \R^m$ for $m = | \text{dep}(i)|$.
We are given intervals $\oin$ such that $\zin_j \in \oin_j = [\lin_j, \uin_j]$ for each coordinate $j$ and want to compute $\oout = [\lout, \uout]$.% based on $g$. 

\paragraph{Softmax layer.}
The softmax function is often used to convert activations into
a probability distribution, e.g., for attention.~\citet{gowal2018effectiveness} uses unnormalized logits and does not handle softmax operations.
Formally, let $\zout$ represent the normalized score of the word at position $c$. We have $\zout = \frac{\exp(\zin_c)}{\sum_{j=1}^m \exp(\zin_j)}$. The value of $\zout$ is largest when $\zin_c$ takes its largest value and all other words take the smallest value:
\begin{align}
  \label{eqn:softmax}
\uout &= \frac{\exp(\uin_c)}{\exp(\uin_c) + \sum \limits_{j\neq c}\exp(\lin_j)}.
\end{align}
We obtain a similar expression for $\lout$. Note that $\lout$ and $\uout$ can each be computed in a forward pass, with some care taken to avoid numerical instability (see \refapp{supp-softmax}).

\paragraph{Element-wise multiplication and dot product.}
Models like LSTMs incorporate gates which perform element-wise multiplication of two activations.
Let $\zout = \zin_1 \zin_2$ where $\zout, \zin_1, \zin_2 \in \R$. 
The extreme values of the product occur at one of the four points
corresponding to the products of the extreme values of the inputs.
In other words,
\begin{align}
  \text{\sC} = \{ \lin_1 \lin_2, ~~~&\lin_1 \uin_2 \nonumber \\
                       \uin_1 \lin_2, ~~~&\uin_1 \uin_2 \Huge \} \nonumber \\
  \lout = \min \big(\sC \big) ~~~~&~~ \uout = \max \big( \sC \big).
\end{align}
Propagating intervals through multiplication nodes therefore requires four multiplications.

Dot products between activations are often used to compute attention scores.\footnote{This is distinct from an affine transformation, because both 
vectors have associated bounds;
in an affine layer, the input has bounds, but the weight matrix is fixed.}
The dot product $(\zin_1)^\top \zin_2$ is just the sum of the element-wise product $\zin_1 \odot \zin_2$. Therefore, we can bound the dot product by summing the bounds on each element of $\zin_1 \odot \zin_2$, using the formula for element-wise multiplication.

\subsection{Final layer}
\label{sec:final-layer}
Classification models typically output a single logit for binary classification, or $k$ logits for $k$-way classification. 
The final loss $f(z, \theta)$ is a function of the logits $s(x)$.
For standard loss functions, we can represent this function in terms of element-wise monotonic functions (\refapp{ibp}) and the elementary functions described in Section~\ref{sec:elem}.
\begin{enumerate}
\item Zero-one loss: $f(z, \theta)=\1[\max (s(x)) = y]$ involves a max operation followed by a step function, which is monotonic.
\item Cross entropy: For multi-class, $f(z, \theta) = \text{softmax}(s(x))$. In the binary case, $f(z, \theta) = \sigma(s(x))$, where the sigmoid function $\sigma$ is monotonic.
\end{enumerate}
Thus, we can compute bounds on the loss $\sO(z, \theta) = [\ell^\text{final}, u^\text{final}]$ from bounds on the logits.

\subsection{Certifiably Robust Training with IBP}
Finally, we describe certifiably robust training, in which we encourage robustness
by minimizing the upper bound on the worst-case loss \cite{dvijotham2018training,gowal2018effectiveness}.
Recall that for an example $z$ and parameters $\theta$,
$u^{\text{final}}(z, \theta)$ is the upper bound on
the loss $f(z, \theta)$.
Given a dataset $D$, we optimize a weighted combination of the normal loss
and the upper bound $u^{\text{final}}$,
\begin{align}
  \label{eqn:final-obj}
    \min_\theta \sum_{z \in D} (1 - \kappa) f(z, \theta) + \kappa \, u^{\text{final}}(z, \theta),
\end{align}
where $0 \le \kappa \le 1$ is a scalar hyperparameter.

As described above, we compute $u^\text{final}$ in a modular fashion: 
each layer has an accompanying function that computes bounds on its outputs given bounds on
its inputs.
Therefore, we can easily apply IBP to new architectures.
Bounds propagate through layers via forward passes,
so the entire objective~\refeqn{final-obj} can be optimized via
backpropagation. 

\citet{gowal2018effectiveness} found that this objective was easier to optimize by
starting with a smaller space of allowed perturbations,
and make it larger during training.
We accomplish this by artificially shrinking the input layer intervals 
$\sO^\text{input}_{ij} = [\ell^\text{input}_{ij}, u^\text{input}_{ij}]$ 
towards the original value $\phi(x_i)_j$
by a factor of $\epsilon$:
\begin{align*}
  \ell^{\text{input}}_{ij} &\leftarrow \phi(x_i)_j - \epsilon (\phi(x_i)_j - \ell^{\text{input}}_{ij}) \\
  u^{\text{input}}_{ij} & \leftarrow \phi(x_i)_j + \epsilon (u^{\text{input}}_{ij} - \phi(x_i)_j).
\end{align*}
Standard training corresponds to $\epsilon=0$.
We train for $T^{\text{init}}$ epochs while linearly increasing 
$\epsilon$ from $0$ to $1$, and also increasing
$\kappa$ from $0$ up to a maximum value of $\kappa^\star$,
We then train for an additional $T^{\text{final}}$ epochs at 
$\kappa = \kappa^\star$ and  $\epsilon = 1$.

To summarize, we use IBP to compute an upper bound on the model's loss when 
given an adversarially perturbed input.
This bound is computed in a modular fashion.
We efficiently train models to minimize this bound via backpropagation.

\section{Tasks and models}
Now we describe the tasks and model architectures on which we run experiments.
These models are all built from the primitives in \refsec{ibp}.

\subsection{Tasks}
Following \citet{alzantot2018adversarial},
we evaluate on two standard NLP datasets:
the IMDB sentiment analysis dataset \citep{maas2011imdb}
and the Stanford Natural Language Inference (SNLI) dataset \citep{bowman2015large}.
For IMDB, the model is given a movie review and must classify it as positive or negative.
For SNLI, the model is given two sentences, a premise and a hypothesis,
and is asked whether the premise entails, contradicts, or is neutral with respect to the hypothesis.
For SNLI, the adversary is only allowed to change the hypothesis,
as in \citet{alzantot2018adversarial},
though it is possible to also allow changing the premise.

\subsection{Models}
\label{sec:models}
\paragraph{IMDB.} We implemented three models for IMDB.
The bag-of-words model (\bowimdb) averages the word vectors
for each word in the input, then passes this through a two-layer feedforward network
with $100$-dimensional hidden state to obtain a final logit.
The other models are similar, except they run either a CNN or bidirectional LSTM
on the word vectors, then average their hidden states.
All models are trained on cross entropy loss.

\paragraph{SNLI} We implemented two models for SNLI.
The bag-of-words model (\bowsnli) encodes
the premise and hypothesis separately by summing their word vectors,
then feeds the concatenation of these encodings to a 3-layer feedforward network.
We also reimplement the Decomposable Attention model \cite{parikh2016decomposable},
which uses attention between the premise and hypothesis 
to compute richer representations of each word in both sentences.
These context-aware vectors are used in the same way \bowsnli uses the original word vectors
to generate the final prediction. 
Both models are trained on cross entropy loss.
Implementation details are provided in \refapp{supp-hyperparams}.

\paragraph{Word vector layer.}
The choice of word vectors affects the tightness of our interval bounds.
We choose to define the word vector $\phi(w)$ for word $w$ as the output of a feedforward layer
applied to a fixed pre-trained word vector $\vpre(w)$:
\begin{align}
  \phi(w) = \relu(\finput(\vpre(w))),
\end{align}
where $\finput$ is a learned linear transformation.
Learning $\finput$ with certifiably robust training encourages it to orient
the word vectors so that the convex hull of the word vectors is close to an axis-aligned box.
Note that $\finput$ is applied \emph{before} bounds are computed via \refeqn{input}.\footnote{
  Equation \refeqn{input} must be applied before the model can 
  combine information from multiple words, but it can be delayed until after
  processing each word independently.}
Applying $\finput$ after the bound calculation would result in looser interval bounds,
since the original word vectors $\vpre(w)$ might be poorly approximated by interval bounds
(e.g., \reffig{transform}a),
compared to $\phi(w)$ (e.g., \reffig{transform}b).
Section~\ref{sec:wordvec} confirms the importance of adding $\finput$.
We use $300$-dimensional GloVe vectors \citep{pennington2014glove} as our $\vpre(w)$.

\section{Experiments}
\label{sec:experiments}
\subsection{Setup}
\label{sec:exp-setup}
\paragraph{Word substitution perturbations.}
We base our sets of allowed word substitutions $S(x, i)$ on 
the substitutions allowed by \citet{alzantot2018adversarial}.
They demonstrated that their substitutions lead to adversarial examples
that are qualitatively similar to the original input and retain the original label,
as judged by humans.
\citet{alzantot2018adversarial} define the neighbors $N(w)$ of a word $w$
as the $n=8$ nearest neighbors of $w$
in a ``counter-fitted'' word vector space
where antonyms are far apart \citep{mrksic2016counterfitting}.\footnote{
  Note that the model itself classifies using a different set of pre-trained word vectors;
  the counter-fitted vectors are only used to define the set of allowed substitution words.}
The neighbors must also lie within some Euclidean distance threshold.
They also use a language model constraint to avoid nonsensical perturbations: 
they allow substituting $x_i$ with $\tilde{x}_i \in N(x_i)$ if and
only if it does not decrease the log-likelihood of the text under a pre-trained language model by more than some threshold.

We make three modifications to this approach.
First, in \citet{alzantot2018adversarial},
the adversary applies substitutions one at a time,
and the neighborhoods and language model scores are computed
relative to the current altered version of the input.
This results in a hard-to-define attack surface,
as changing one word can allow or disallow changes to other words.
It also requires recomputing language model scores at each iteration of the genetic attack,
which is inefficient.
Moreover, the same word can be substituted multiple times, leading to semantic drift.
We define allowed substitutions relative to the original sentence $x$,
and disallow repeated substitutions.
Second, we use a faster language model that allows us to query longer contexts;
\citet{alzantot2018adversarial} use a slower language model
and could only query it with short contexts.
Finally, we use the language model constraint only at test time;
the model is trained against all perturbations in $N(w)$.
This encourages the model to be robust to 
a larger space of perturbations, instead of specializing for the particular
choice of language model.
See \refapp{supp-attack} for further details.

\paragraph{Analysis of word neighbors.}
One natural question is whether we could guarantee robustness
by having the model treat all neighboring words the same.
We could construct equivalence classes of words from the transitive closure of $N(w)$,
and represent each equivalence class with one embedding.
We found that this would lose a significant amount of information.
Out of the 50,000 word vocabulary,
19,122 words would be in the same equivalence class,
including the words ``good'',  ``bad'', ``excellent'', and ``terrible.''
Of the remaining words, 24,389 ($79\%$) have no neighbors.

\paragraph{Baseline training methods.}
We compare certifiably robust training (\refsec{ibp}) with both 
standard training and data augmentation,
which has been used in NLP to encourage robustness to various types of perturbations \cite{jia2017adversarial,belinkov2017synthetic,iyyer2018adversarial,ribeiro2018sears}.
In data augmentation, for each training example $z$, we augment the dataset with $K$ new examples
$\tilde{z}$ by sampling $\tilde{z}$ uniformly from $\ball{z}$,
then train on the normal cross entropy loss.
For our main experiments, we use $K=4$.
We do not use adversarial training \cite{goodfellow2015explaining}
because it would require running an adversarial search procedure at each training step,
which would be prohibitively slow.

\paragraph{Evaluation of robustness.}
We wish to evaluate robustness of models to all word substitution perturbations.
Ideally, we would directly measure \emph{robust accuracy}, the fraction of
test examples $z$ for which the model is correct on all $\tilde{z} \in \ball{z}$.
However, evaluating this exactly involves enumerating the exponentially large set of perturbations, which is intractable.
Instead, we compute tractable upper and lower bounds:
\begin{enumerate}[1.]
\item Genetic attack accuracy: \citet{alzantot2018adversarial} 
  demonstrate the effectiveness of a genetic algorithm that searches for perturbations $\tilde{z}$ that cause model misclassification.
  The algorithm maintains a ``population'' of candidate $\tilde{z}$'s
  and repeatedly perturbs and combines them. 
  We used a population size of $60$ and ran $40$ search iterations on each example.
  Since the algorithm does not exhaustively search over $\ball{z}$,
  accuracy on the perturbations it finds is an \emph{upper bound} on the true robust accuracy.
\item Certified accuracy: To complement this upper bound,
  we use IBP to obtain a tractable lower bound on the robust accuracy.
  Recall from \refsec{final-layer} that we can use IBP to get an upper bound on the zero-one loss.
  From this, we obtain a \emph{lower bound} on the robust accuracy by measuring
  the fraction of test examples for which the zero-one loss is guaranteed to be $0$.
\end{enumerate}

\paragraph{Experimental details.}
For IMDB, we split the official train set into train and development subsets,
putting reviews for different movies into different splits
(matching the original train/test split).
For SNLI, we use the official train/development/test split.
We tune hyperparameters on the development set for each dataset.
Hyperparameters are reported in \refapp{supp-hyperparams}.

\subsection{Main results}
\label{sec:results}
\begin{table}[t]
  \small
  \centering
  \begin{tabular}{|l|c|c|}
    \hline
    \multirow{2}{*}{System} & \stackunder{Genetic attack}{(Upper bound)} & \stackunder{IBP-certified}{(Lower bound)} \\
    \hline
    \textbf{Standard training} & & \\
    \bowimdb      &$ 9.6$ & $ 0.8$ \\
    CNN           &$ 7.9$ & $ 0.1$ \\
    LSTM          &$ 6.9$ & $ 0.0$ \\
    \hline
    \textbf{Robust training} & & \\
    \bowimdb      & $70.5$ & $68.9$ \\
    CNN           & $\bf 75.0$ & $\bf 74.2$ \\
    LSTM          & $64.7$ & $63.0$ \\
    \hline
    \textbf{Data augmentation} & & \\
    \bowimdb      & $34.6$ & $ 3.5$ \\
    CNN           & $35.2$ & $ 0.3$ \\
    LSTM          & $33.0$ & $ 0.0$ \\
    \hline
  \end{tabular}
  \caption{Robustness of models on IMDB. We report accuracy on perturbations obtained via the genetic attack (upper bound on robust accuracy), and certified accuracy obtained using IBP (lower bound on robust accuracy) on $1000$ random IMDB test set examples. For all models, robust training vastly outperforms data augmentation 
  ($p<10^{-63}$, Wilcoxon signed-rank test).
  }
  \label{tab:imdb}
\end{table}

\begin{table}[t]
  \small
  \centering
  \begin{tabular}{|l|c|c|}
    \hline
    \multirow{2}{*}{System} & \stackunder{Genetic attack}{(Upper bound)} & \stackunder{IBP-certified}{(Lower bound)} \\
    \hline
    \textbf{Normal training} & & \\
    \bowsnli      & $40.5$ & $2.3$ \\
    \decompattn   & $40.3$ & $1.4$ \\
    \hline
    \textbf{Robust training} & & \\
    \bowsnli      & $\bf 75.0$ & $\bf 72.7$ \\
    \decompattn   & $73.7$ & $72.4$ \\
    \hline
    \textbf{Data augmentation} & & \\
    \bowsnli      & $68.5$ & $7.7$ \\
    \decompattn   & $70.8$ & $1.4$ \\
    \hline
  \end{tabular}
  \caption{
  Robustness of models on the SNLI test set.
  For both models, robust training outperforms data augmentation
  ($p<10^{-10}$, Wilcoxon signed-rank test).}
  \label{tab:snli}
\end{table}

\reftab{imdb} and \reftab{snli} show our main results for IMDB and SNLI, respectively.
We measure accuracy on perturbations found by the genetic attack (upper bound on robust accuracy)
and IBP-certified accuracy (lower bound on robust accuracy) on $1000$ random test examples from IMDB,\footnote{We downsample the test set because the genetic attack is slow on IMDB,
as inputs can be hundreds of words long.}
and all $9824$ test examples from SNLI.
Across many architectures, 
our models are more robust to perturbations than ones trained with data augmentation.
This effect is especially pronounced on IMDB, where inputs can be hundreds of words long, so many words can be perturbed.
On IMDB, the best IBP-trained model gets $75.0\%$ accuracy 
on perturbations found by the genetic attack, 
whereas the best data augmentation model gets $35.2\%$.
Normally trained models are even worse, with adversarial accuracies below $10\%$.

\paragraph{Certified accuracy.}
Certifiably robust training yields models with tight guarantees on robustness---the upper and lower bounds on robust accuracy are close. 
On IMDB, the best model is \emph{guaranteed} to be correct on all perturbations of $74.2\%$ of test examples,
very close to the $75.0\%$ accuracy against the genetic attack.
In contrast, for data augmentation models, the IBP bound cannot guarantee robustness on almost all examples. 
It is possible that a stronger attack (e.g., exhaustive search) could further lower the accuracy of these models, or that the IBP bounds are loose.

LSTM models can be certified with IBP, though they fare worse than other models.
IBP bounds may be loose for RNNs
because of their long computation paths, along which looseness of bounds can get amplified.
Nonetheless, in \refapp{lstm}, we show on synthetic data that robustly trained LSTMs can learn long-range dependencies.

\subsection{Clean versus robust accuracy}
\label{sec:tradeoff}
\begin{figure}
  \center
  \includegraphics[width=\linewidth]{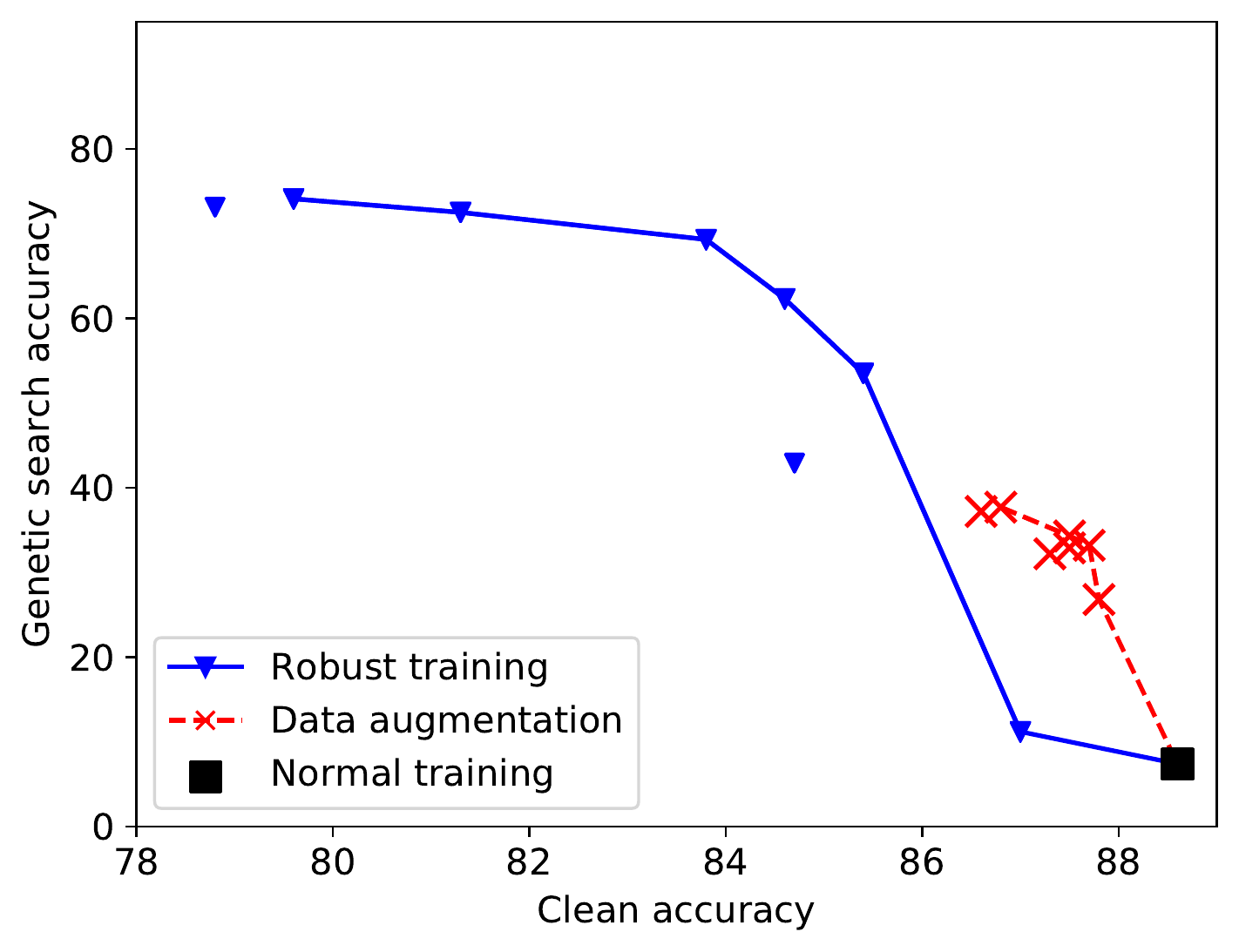}
  \caption{
  Trade-off between clean accuracy and genetic attack accuracy for CNN models on IMDB.
  Data augmentation cannot achieve high robustness. 
  Certifiably robust training yields much more robust models, though at the cost of some clean accuracy.
  Lines connect Pareto optimal points for each training strategy.
  }
  \label{fig:tradeoff}
\end{figure}

Robust training does cause a moderate drop in clean accuracy (accuracy on unperturbed test examples)
compared with normal training. 
On IMDB, our normally trained CNN model gets $89\%$ clean accuracy, compared to $81\%$ for the robustly trained model.
We also see a drop on SNLI: the normally trained \bowsnli model gets $83\%$ clean accuracy, compared to $79\%$ for the robustly trained model.
Similar drops in clean accuracy are also seen for robust models in vision \cite{madry2017towards}. For example, the state-of-the-art robust model on CIFAR10 \cite{zhang2019theoretically}
only has $85\%$ clean accuracy, but comparable normally-trained models get $>96\%$ accuracy. 

We found that the robustly trained models tend to underfit the training data---on IMDB, the CNN model gets only $86\%$ clean training accuracy, 
lower than the \emph{test} accuracy of the normally trained model.
The model continued to underfit when we increased either the depth or width of the network.
One possible explanation is that the attack surface adds a lot of noise,
though a large enough model should still be able to overfit the training set.
Better optimization or a tighter way to compute bounds could also improve training accuracy.
We leave further exploration to future work.

Next, we analyzed the trade-off between clean and robust accuracy
by varying the importance placed on perturbed examples during training.
We use accuracy against the genetic attack as our proxy for robust accuracy,
rather than IBP-certified accuracy, as IBP bounds may be loose
for models that were not trained with IBP.
For data augmentation, we vary $K$, the number of augmented examples per real example, from $1$ to $64$.
For certifiably robust training, we vary $\kappa^\star$, the weight of the certified robustness training objective, between $0.01$ and $1.0$.
\reffig{tradeoff} shows trade-off curves for the CNN model on $1000$ random IMDB development set examples.
Data augmentation can increase robustness somewhat,
but cannot reach very high adversarial accuracy.
With certifiably robust training, we can trade off some clean accuracy for much higher robust accuracy.

\subsection{Runtime considerations}
IBP enables efficient computation of $u^{\text{final}}(z, \theta)$,
but it still incurs some overhead.
Across model architectures, we found that one epoch of certifiably robust training takes between $2\times$ and $4\times$ longer than one epoch of standard training.
On the other hand, IBP certificates are much faster to compute at test time than 
genetic attack accuracy.
For the robustly trained CNN IMDB model,
computing certificates on $1000$ test examples took 5 seconds,
while running the genetic attack on those same examples took over 3 hours.

\subsection{Error analysis}

\begin{figure}
  \center
  \includegraphics[width=\linewidth]{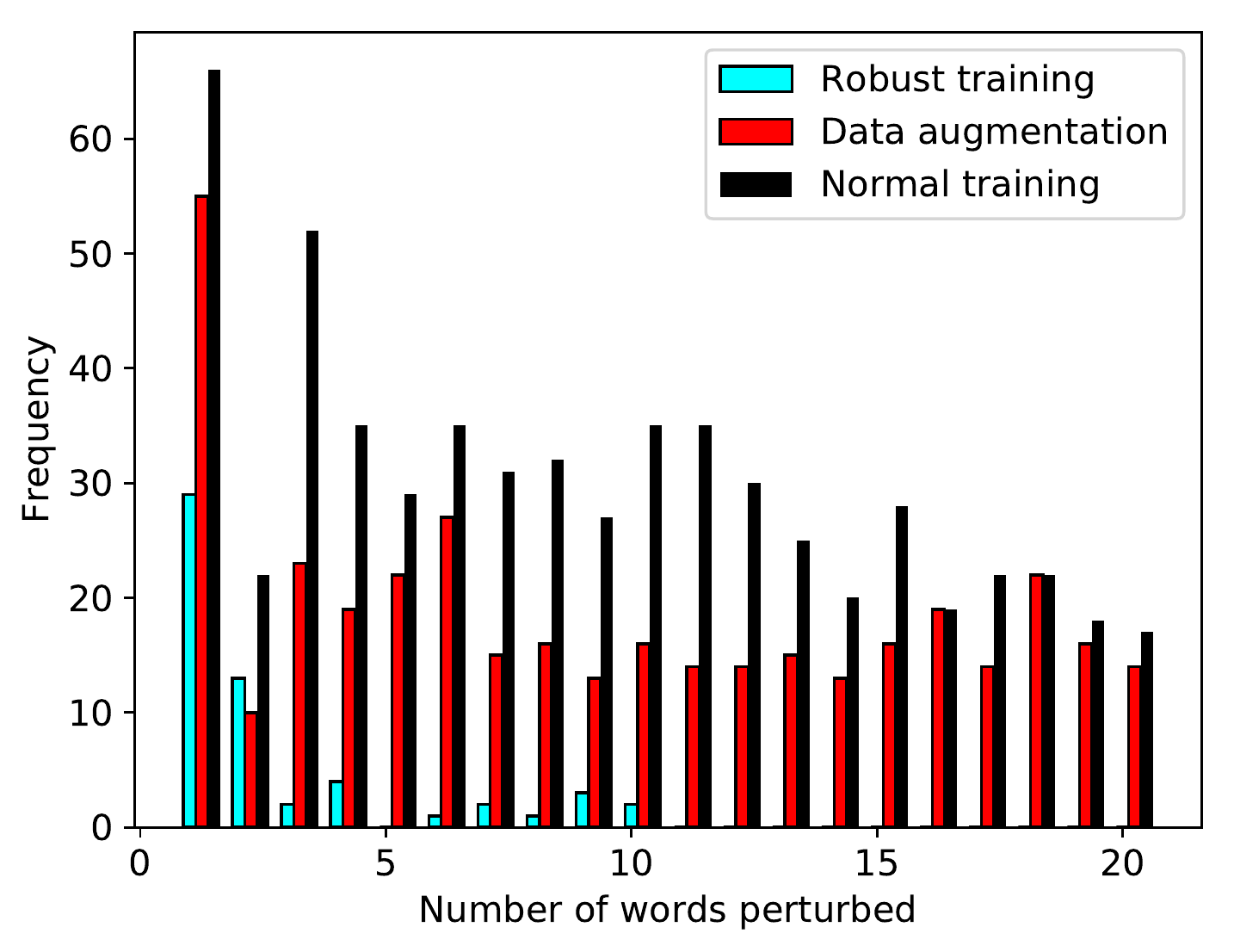}
  \caption{
    Number of words perturbed by the genetic attack
    to cause errors by CNN models
    on $1000$ IMDB development set examples.
    Certifiably robust training reduces the effect of many simultaneous perturbations.
  }
  \label{fig:perturb-count}
\end{figure}
We examined development set examples on which models were correct on the original input
but incorrect on the perturbation found by the genetic attack.
We refer to such cases as \emph{robustness errors}.
We focused on the CNN IMDB models trained normally, robustly, and with data augmentation.
We found that robustness errors of the robustly trained model mostly occurred when it was
not confident in its original prediction.
The model had $>70\%$ confidence in the correct class for the original input
in only $14\%$ of robustness errors.
In contrast, the normally trained and data augmentation models were more confident on their robustness errors;
they had $>70\%$ confidence on the original example
in $92\%$ and $87\%$ of cases, respectively.

We next investigated how many words the genetic attack needed to change to cause misclassification,
as shown in \reffig{perturb-count}.
For the normally trained model, some robustness errors involved only a couple changed words 
(e.g., \nl{I've finally found a movie worse than \dots} was classified negative,
but the same review with \nl{I've finally \textbf{discovered} a movie worse than\dots} 
was classified positive),
but more changes were also common
(e.g., part of a review was changed from \nl{The creature looked very cheesy} to
\nl{The creature \textbf{seemed supremely dorky}},
with $15$ words changed in total).
Surprisingly, certifiably robust training nearly eliminated robustness errors in which the genetic attack had to change many words:
the genetic attack either caused an error by changing a couple words,
or was unable to trigger an error at all.
In contrast, data augmentation is unable to cover the exponentially large space of perturbations that involve many words, so it does not prevent errors caused by changing many words.

\subsection{Training schedule}
\label{sec:training}
We investigated the importance of slowly increasing $\epsilon$ during training,
as suggested by \citet{gowal2018effectiveness}.
Fixing $\epsilon=1$ during training led to a $5$ point reduction in certified accuracy for the CNN.
On the other hand, we found that holding $\kappa$ fixed did not hurt accuracy,
and in fact may be preferable.
More details are shown in \refapp{supp-training}.

\subsection{Word vector analysis}
\label{sec:wordvec}
We determined the importance of the extra feedforward layer $\finput$
that we apply to pre-trained word vectors, as described in \refsec{models}.
We compared with directly using pre-trained word vectors, i.e.
$\phi(w) = \vpre(w)$.
We also tried using $\finput$ but applying interval bounds on $\vpre(w)$,
then computing bounds on $\phi(w)$ with the IBP formula for affine layers.
In both cases, we could not train a CNN to achieve more than $52.2\%$ certified accuracy
on the development set.
Thus, transforming pre-trained word vectors and applying interval bounds \emph{after}
is crucial for robust training.
In \refapp{supp-wordvec},
we show that robust training makes the intervals around 
transformed word vectors smaller, compared to the pre-trained vectors.

\section{Related Work and Discussion}
Recent work on adversarial examples in NLP has proposed
various classes of perturbations, such as
insertion of extraneous text \cite{jia2017adversarial},
word substitutions \cite{alzantot2018adversarial},
paraphrasing \cite{iyyer2018adversarial,ribeiro2018sears},
and character-level noise \cite{belinkov2017synthetic, ebrahimi2017hotflip}.
These works focus mainly on demonstrating models' lack of robustness,
and mostly do not explore ways to increase robustness beyond data augmentation.
Data augmentation is effective for narrow perturbation spaces
\cite{jia2017adversarial,ribeiro2018sears},
but only confers partial robustness in other cases \cite{iyyer2018adversarial,alzantot2018adversarial}.
\citet{ebrahimi2017hotflip} tried adversarial training \cite{goodfellow2015explaining} for character-level perturbations,
but could only use a fast heuristic attack at training time, due to runtime considerations.
As a result, their models were still be fooled by running a more expensive search procedure at test time.

Provable defenses have been studied for simpler NLP models and attacks,
particularly for tasks like spam detection
where real-life adversaries try to evade detection.
\citet{globerson2006nightmare} train linear classifiers that are robust to
adversarial feature deletion.
\citet{dalvi2004adversarial} 
analyzed optimal strategies for a Naive Bayes classifier and attacker,
but their classifier only defends against a fixed attacker that does not
adapt to the model.

Recent work in computer vision
\citep{szegedy2014intriguing,goodfellow2015explaining}
has sparked renewed interest in adversarial examples.
Most work in this area focuses on $L_\infty$-bounded perturbations, 
in which each input pixel can be changed by a small amount.
The word substitution attack model we consider is similar to $L_\infty$ perturbations,
as the adversary can change each input word by a small amount.
Our work is inspired by
work based on convex optimization \cite{raghunathan2018certified,wong2018provable}
and builds directly on interval bound propagation \cite{dvijotham2018training,gowal2018effectiveness},
which has certified robustness of computer vision models to $L_\infty$ attacks.
Adversarial training via projected gradient descent \citep{madry2018towards} 
has also been shown to improve robustness, but assumes that inputs are continuous.
It could be applied in NLP by
relaxing sets of word vectors to continuous regions.

This work provides certificates against word substitution perturbations for particular models.
Since IBP is modular, it can be extended to other model architectures on other tasks.
It is an open question whether IBP can give non-trivial bounds
for sequence-to-sequence tasks like machine translation \citep{belinkov2017synthetic,michel2019adversarial}.
In principle, IBP can handle character-level typos \cite{ebrahimi2017hotflip,pruthi2019misspellings},
though typos yield more perturbations per word than we consider in this work.
We are also interested in handling word insertions and deletions, rather than just substitutions.
Finally, we would like to train models that get state-of-the-art clean accuracy while also being provably robust; achieving this remains an open problem.

In conclusion, state-of-the-art NLP models are accurate on average,
but they still have significant blind spots.
Certifiably robust training provides a general, principled mechanism to 
avoid such blind spots by encouraging models to make correct predictions on
all inputs within some known perturbation neighborhood.
This type of robustness is a necessary (but not sufficient) property of models that truly understand language.
We hope that our work is a stepping stone towards models that are robust against an even wider, harder-to-characterize space of possible attacks.

\section*{Acknowledgments}
This work was supported by NSF Award Grant no. 1805310
and the DARPA ASED program under FA8650-18-2-7882.
R.J. is supported by an NSF Graduate Research Fellowship under Grant No. DGE-114747.
A.R. is supported by a Google PhD Fellowship and the Open Philanthropy Project AI Fellowship. We thank Allen Nie for providing the pre-trained language model,
and thank Peng Qi, Urvashi Khandelwal, Shiori Sagawa, and the anonymous reviewers for their helpful comments.

\section*{Reproducibility} All code, data, and experiments are available on Codalab at 
\url{https://bit.ly/2KVxIFN}.

\bibliography{refdb/all}
\bibliographystyle{acl_natbib}

\appendix
\section{Supplemental material}
\label{sec:supplement}

\subsection{Additional interval bound formulas}
\label{sec:supp-ibp}
\citet{gowal2018effectiveness} showed how to compute interval bounds for affine transformations
and monotonic element-wise functions. Here, we review their derivations, for completeness.

\paragraph{Affine transformations.}
Affine transformations are the building blocks of neural networks. Suppose $\zout = a^\top \zin + b$ for weight $a \in \R^m$ and bias $b \in \R$. $\zout$ is largest when positive entries of $a$ are multiplied with $\uin$ and negative with $\lin$:
\begin{align}
  \label{eqn:affine}
  \uout &= \underbrace{0.5(a + |a|)^\top}_\text{positive} \uin + \underbrace{0.5(a - |a|)^\top}_\text{negative} \lin + b\nonumber \\
  &= \mu + r,
\end{align}
where $\mu = 0.5 a^\top(\lin + \uin) + b$ and $r = 0.5 |a|^\top(u - l)$. A similar computation yields that $\lout = \mu - r$.
Therefore, the interval $\oout$ can be computed using two inner product evaluations: one with $a$
and one with $|a|$. 

\paragraph{Monotonic scalar functions.}
Activation functions such as ReLU, sigmoid and tanh are monotonic.
Suppose $\zout = \sigma(\zin)$ where $\zout, \zin \in \R$,
i.e. the node applies an element-wise function to its input.
The intervals can be computed trivially since $\zout$ is minimized at $\lin$
and maximized at $\uin$.
\begin{align}
    \label{eqn:activation}
    \lout = \sigma(\lin), ~~ \uout = \sigma(\uin).
\end{align}

\subsection{Numerical stability of softmax}
\label{sec:supp-softmax}
\newcommand{\logsumexp}{\operatorname{logsumexp}}
\newcommand{\logonep}{\operatorname{log1p}}
\newcommand{\expmone}{\operatorname{expm1}}
In this section, we show how to compute interval bounds for softmax layers
in a numerically stable way.
We will do this by showing how to handle log-softmax layers.
Note that since softmax is just exponentiated log-softmax,
and exponentiation is monotonic, bounds on log-softmax directly yield bounds on softmax.

Let $\zin$ denote a vector of length $m$, let $c$ be an integer $\in \{1, \dotsc, m\}$,
and let $\zout$ represent the log-softmax score of index $c$, i.e. 
\begin{align}
\zout
&= \log \frac{\exp(\zin_c)}{\sum_{j=1}^m \exp(\zin_j)} \\
&= \zin_c - \log \sum_{j=1}^m \exp(\zin_j).
\end{align}
Given interval bounds $\ell_j \le \zin_j \le u_j$ for each $j$,
we show how to compute upper and lower bounds on $\zout$.
For any vector $v$, we assume access to a subroutine that computes \[
\logsumexp(v) = \log \sum_i \exp(v_i)
\]
stably. 
The standard way to compute this is to normalize $v$ by subtracting $\max_i(v_i)$ before taking exponentials,
then add it back at the end.
$\logsumexp$ is a standard function in libraries like PyTorch.
We will also rely on the 
fact that if $v$ is the concatenation of vectors $u$ and $w$, then 
$\logsumexp(v) = \logsumexp([\logsumexp(u), \logsumexp(w)])$.

\paragraph{Upper bound.}
The upper bound $\uout$ is achieved by having the maximum value of $\zin_c$,
and minimum value of all others. This can be written as:

{\small
\begin{align}
\uout 
&= \uin_c - \log\left(\exp(\uin_c) + \sum_{1 \le j \le m, j \ne c} \exp(\lin_j) \right).
\end{align}
}%
While we could directly compute this expression, it is difficult to vectorize.
Instead, with some rearranging, we get

{\small
\begin{align}
\uout &= \uin_c - \log\left(\exp(\uin_c) - \exp(\lin_c) + \sum_{j=1}^m \exp(\lin_j)\right).
\end{align}
}%
The second term is the $\logsumexp$ of 
\begin{align}
  \label{eqn:exp-diff}
\log\left(\exp(\uin_c) - \exp(\lin_c)\right)
\end{align} 
and 
\begin{align}
\logsumexp(\lin).
\end{align}
Since we know how to compute $\logsumexp$, this reduces to computing \refeqn{exp-diff}.
Note that \refeqn{exp-diff} can be rewritten as
\begin{align}
\uin_c + \log\left(1 - \exp(\lin_c - \uin_c)\right)
\end{align}
by adding and subtracting $\uin_c$.
To compute this quantity, we consider two cases:
\begin{enumerate}
\item $\uin_c \gg \lin_c$. 
Here we use the fact that stable methods exist to compute $\logonep(x) = \log(1 + x)$ for $x$ close to $0$.
We compute the desired value as \[
\uin_c + \log1p(-\exp(\lin_c - \uin_c)),
\]
since $\exp(\lin_c - \uin_c)$ will be close to $0$.
\item $\uin_c$ close to $\lin_c$. 
Here we use the fact that stable methods exist to compute $\expmone(x) = \exp(x) - 1$ for $x$ close to $0$.
We compute the desired value as \[
\uin_c + \log(-\expmone(\lin_c - \uin_c)),
\]
since $\lin_c - \uin_c$ may be close to $0$.
\end{enumerate}
We use case 1 if $\uin_c - \lin_c > \log2$, and case 2 otherwise.\footnote{See \url{https://cran.r-project.org/web/packages/Rmpfr/vignettes/log1mexp-note.pdf} for further explanation.}

\paragraph{Lower bound.}
The lower bound $\lout$ is achieved by having the minimum value of $\zin_c$,
and the maximum value of all others. This can be written as:

{\small
\begin{align}
\lout 
&= \lin_c - \log\left(\exp(\lin_c) + \sum_{1 \le j \le m, j \ne c} \exp(\uin_j) \right).
\end{align}
}%
The second term is just a normal $\logsumexp$, which is easy to compute.
To vectorize the implementation, it helps to first compute
the $\logsumexp$ of everything except $\lin_c$, and then $\logsumexp$ that with $\lin_c$.

\subsection{Attack surface differences}
\label{sec:supp-attack}

In \citet{alzantot2018adversarial}, 
the adversary applies replacements one at a time,
and the neighborhoods and language model scores are computed
relative to the current altered version of the input.
This results in a hard-to-define attack surface,
as the same word can be replaced many times, leading to semantic drift.
We instead pre-compute the allowed substitutions $S(x, i)$ at index $i$
based on the original $x$. We define $S(x, i)$ as the set of
$\tilde{x}_i \in N(x_i)$ such that
\begin{align}
  \log & P(x_{i-W:i-1}, \tilde{x}_i, x_{i+1:i+W}) \ge \nonumber \\
      & \log P(x_{i-W:i+W}) - \delta
\end{align}
where probabilities are assigned by a pre-trained language model,
and the window radius $W$ and threshold $\delta$ are hyperparameters.
We use $W=6$ and $\delta=5$.
We also use a different language model\footnote{\url{https://github.com/windweller/l2w}} from \citet{alzantot2018adversarial}
that achieves perplexity of $50.79$ on the One Billion Word dataset \citep{chelba2013one}.
\citet{alzantot2018adversarial} use a different, slower language model,
which compels them to use a smaller window radius of $W=1$.

\subsection{Experimental details}
\label{sec:supp-hyperparams}
\begin{table*}[!ht]
  \small
  \centering
  \begin{tabular}{|l|cccccc|}
    \hline
    System & $\kappa$ & Learning Rate & Dropout Prob. & Weight Decay & Gradient Norm Clip Val. & $T^{init}$ \\
    \hline
    IMDB, BOW          & $0.8$ & $1\times 10^{-3}$ & $0.2$ & $1\times 10^{-4}$ & $0.25$ & $40$ \\
    IMDB, CNN          & $0.8$ & $1\times 10^{-3}$ & $0.2$ & $1\times 10^{-4}$ & $0.25$ & $40$ \\
    IMDB, LSTM         & $0.8$ & $1\times 10^{-3}$ & $0.2$ & $1\times 10^{-4}$ & $0.25$ & $20$ \\
    SNLI, \bowsnli     & $0.5$ & $5\times 10^{-4}$ & $0.1$ & $1\times 10^{-4}$ & $0.25$ & $35$ \\
    SNLI, \decompattn  & $0.5$ & $1\times 10^{-4}$ & $0.1$ & $0$    & $0.25$ & $50$ \\
    \hline
  \end{tabular}
  \caption{Training hyperparameters for training the models.
  The same hyperparameters were used for all training settings(plain, data augmentation, robust training)
  }
\end{table*}

We do not run training for a set number of epochs but do early stopping on the development set instead.
For normal training, we early stop on normal development set accuracy.
For training with data augmentation, we early stop on the accuracy on the augmented development set.
For certifiably robust training, we early stop on the certifiably robust accuracy on the development set.
We use the Adam optimizer \citep{kingma2014adam} to train all models.

On IMDB, we restrict the model to only use the $50,000$ words
that are in the vocabulary of the counter-fitted word vector space
of \citet{mrksic2016counterfitting}.
This is because perturbations are not allowed for any words not in this vocabulary,
i.e. $N(w) = \{w\}$ for $w \notin V$.
Therefore, the model is strongly incentivized to predict based on
words outside of this set.
While this is a valid way to achieve high certified accuracy, 
it is not a valid robustness strategy in general.
We simply delete all words that are not in the vocabulary before feeding the input to the model.

For SNLI, we use $100$-dimensional hidden state for the \bowsnli model and a $3$-layer feedforward network.
These values were chosen by a hyperparameter search on the dev set.
For \decompattn, we use a $300$-dimensional hidden state and a $2$-layer feedforward network on top of the context-aware vectors.
These values were chosen to match \citet{parikh2016decomposable}.

Our implementation of the Decomposable Attention follows the original described in \cite{parikh2016decomposable}
except for a few differences listed below;
\begin{itemize}
  \item We do not normalize GloVe vectors to have norm 1.
  \item We do not hash out-of-vocabulary words to randomly generated vectors that we train, instead we omit them.
  \item We do randomly generate a null token vector that we then train. (Whether the null vector is trained is unspecified in the original paper).
  \item We use the Adam optimizer (with a learning rate of $1\times 10^{-4}$) instead of AdaGrad.
  \item We use a batch size of 256 instead of 4.
  \item We use a dropout probability of $0.1$ instead of $0.2$
  \item We do not use the intra-sentence attention module.
\end{itemize}

\subsection{Training schedule}
\label{sec:supp-training}
In \reftab{training}, 
we show the effect of holding $\epsilon$ or $\kappa$ fixed during training,
as described in \refsec{training}.
All numbers are on $1000$ randomly chosen examples from the IMDB development set.
Slowly increasing $\epsilon$ is important for good performance.
Slowly increasing $\kappa$ is actually slightly worse than holding $\kappa = \kappa^*$ fixed during training,
despite earlier experiments we ran suggesting the opposite.
Here we only report certified accuracy, as all models are trained with certifiably robust training,
and certified accuracy is much faster to compute for development purposes.

\begin{table}[h]
  \centering
  \begin{tabular}{|l|c|}
    \hline
    \multirow{2}{*}{System} & \stackunder{IBP-certified}{(Lower bound)} \\
    \hline
    BOW                                         & $68.8$ \\
    $\rightarrow$ Fixed $\epsilon$              & $46.6$ \\
    $\rightarrow$ Fixed $\kappa$                & $69.8$ \\
    $\rightarrow$ Fixed $\epsilon$ and $\kappa$ & $66.3$ \\
    \hline
    CNN                                         & $72.5$ \\
    $\rightarrow$ Fixed $\epsilon$              & $67.6$ \\
    $\rightarrow$ Fixed $\kappa$                & $74.5$ \\
    $\rightarrow$ Fixed $\epsilon$ and $\kappa$ & $65.3$ \\
    \hline

    LSTM                                        & $62.5$ \\
    $\rightarrow$ Fixed $\epsilon$              & $43.7$ \\
    $\rightarrow$ Fixed $\kappa$                & $63.0$ \\
    $\rightarrow$ Fixed $\epsilon$ and $\kappa$ & $62.0$ \\
    \hline
  \end{tabular}
  \caption{Effects of holding $\epsilon$ and $\kappa$ fixed during training.
  All numbers are on $1000$ randomly chosen IMDB development set examples.}
  \label{tab:training}
\end{table}

\subsection{Word vector bound sizes}
\label{sec:supp-wordvec}
To better understand the effect of $\finput$,
we checked whether $\finput$ made interval bound boxes
around neighborhoods $N(w)$ smaller.
For each word $w$ with $|N(w)| > 1$, and for both the pre-trained vectors
$\vpre(\cdot)$ and transformed vectors $\phi(\cdot)$,
we compute
\begin{align*}
  \frac1d \sum_{i=1}^d \frac1\sigma_i \left(u^{\text{word}}_w - \ell^{\text{word}}_w\right)
\end{align*}
where $\ell^{\text{word}}_w$ and $u^{\text{word}}_w$ are the interval bounds around 
either $\{\vpre(\tilde{w}): \tilde{w} \in N(w)\}$
or $\{\phi(\tilde{w}): \tilde{w} \in N(w)\}$,
and $\sigma_i$ is the standard deviation across the vocabulary of the $i$-th coordinate of the embeddings. 
This quantity measures the average width of the IBP bounds for the word vectors of $w$ and its neighbors,
normalized by the standard deviation in each coordinate.
On $78.2\%$ of words with $|N(w)| > 1$, this value was smaller for the transformed vectors 
learned by the CNN on IMDB with robust training, compared to the GloVe vectors.
For same model with normal training, the value was smaller only $54.5\%$ of the time,
implying that robust training makes the transformation produce tighter bounds.
We observed the same pattern for other model architectures as well.

\subsection{Certifying long-term memory}
\label{sec:lstm}
We might expect that LSTMs are difficult to certify with IBP,
due to their long computation paths.
To test whether robust training can learn recurrent models
that track state across many time steps,
we created a toy binary classification task
where the input is a sequence of words $x_1, \dotsc, x_L$,
and the label $y$ is $1$ if $x_1 = x_L$ and $0$ otherwise.
We trained an LSTM model that reads the input left-to-right, and tries to predict $y$
with a two-layer feedforward network on top of the final hidden state.  
To do this task, the model must encode the first word in its state
and remember it until the final timestep;
a bag of words model cannot do this task.
For perturbations, we allow replacing
every middle word $x_2, \dotsc, x_{L-1}$ with any word in the vocabulary.
We use robust training on $4000$ randomly generated examples,
where the length of each example is sampled uniformly between $3$ and $10$.
The model obtains $100\%$ certified accuracy on a test set of $1000$ examples,
confirming that robust training can learn models that track state across many time steps.

For this experiment, we found it important to first train for multiple epochs with no certified objective,
before increasing $\epsilon$ and $\kappa$. 
Otherwise, the model gets stuck in bad local optima.
We trained for $50$ epochs using the normal objective, $50$ epochs increasing $\epsilon$ towards $1$ and $\kappa$ towards $0.5$, then $17$ final epochs (determined by early stopping) with these final values of $\epsilon$ and $\kappa$.\footnote{
Note that this dataset is much smaller than IMDB and SNLI, so each epoch corresponds to many fewer parameter updates.}
We leave further exploration of these learning schedule tactics to future work.
We also found it necessary to use a larger LSTM---we used one with $300$-dimensional hidden states.

\clearpage
\newpage
\section{Adversarial examples}
\label{sec:examples}
In this additional supplementary material, we show randomly chosen adversarial examples found by the genetic attack. 
We show examples for three different models: the CNN model on IMDB trained normally, with certifiably robust training, and with data augmentation.
For each model, we picked ten random development set examples for which the model was correct on the original example, but wrong after the genetic attack.
Changed words are marked in bold.

\begin{figure*}
Normally trained model, example 1 \\
\fbox{ \begin{minipage}{\textwidth}
Original: The original is a relaxing watch , with some truly memorable animated sequences . Unfortunately , the sequel , while not the worst of the DTV sequels completely lacks the sparkle . The \textbf{biggest} letdown is a lack of a story . Like Belle 's Magical World , the characters are told through a series of vignettes . Magical World , while marginally \textbf{better} , still manages to make a mess of the story . In between the vignettes , we see the mice at work , and I personally think the antics of Jaq and Gus are the redeeming merits of this movie . The first vignette is the \textbf{best} , about Cinderella getting used to being to being a princess . This is the best , because the mice were at their funniest here . The \textbf{worst} of the vignettes , when Jaq turns into a human , is cute at times , but \textbf{has} a lack of imagination . The last vignette , when Anastasia falls in love , was also cute . The problem was , I could n't imagine Anastasia being friendly with Cinderella , as I considered her the meaner out of the stepsisters . This was also \textbf{marred} by a rather ridiculous subplot about Lucifer falling in love with PomPom . The incidental music was very pleasant to listen to ; however I hated the songs , they were really uninspired , and nothing like the beautiful Tchaikovsky inspired melodies of the original . The characters were the strongest development here . Cinderella while still caring , had lost her sincerity , and a lot of her charm from the original , though she does wear some very pretty clothes . The Duke had some truly funny moments but they were n't enough to save the \textbf{film} , likewise with Prudence and the king . As I mentioned , the mice were the redeeming merits of the movie , as they alone contributed to the film 's cuteness . I have to say also the animation is colourful and above average , and the voice acting was surprisingly good . All in all , a cute , if unoriginal sequel , that was \textbf{marred} by the songs and a lack of a story . 4/10 for the mice , the \textbf{voice} acting , the animation and some pretty dresses . Bethany Cox
\end{minipage} }
\fbox{ \begin{minipage}{\textwidth}
Perturbed: The original is a relaxing watch , with some truly memorable animated sequences . Unfortunately , the sequel , while not the worst of the DTV sequels completely lacks the sparkle . The \textbf{greatest} letdown is a lack of a story . Like Belle 's Magical World , the characters are told through a series of vignettes . Magical World , while marginally \textbf{nicer} , still manages to make a mess of the story . In between the vignettes , we see the mice at work , and I personally think the antics of Jaq and Gus are the redeeming merits of this movie . The first vignette is the \textbf{finest} , about Cinderella getting used to being to being a princess . This is the best , because the mice were at their funniest here . The \textbf{toughest} of the vignettes , when Jaq turns into a human , is cute at times , but \textbf{possesses} a lack of imagination . The last vignette , when Anastasia falls in love , was also cute . The problem was , I could n't imagine Anastasia being friendly with Cinderella , as I considered her the meaner out of the stepsisters . This was also \textbf{tempered} by a rather ridiculous subplot about Lucifer falling in love with PomPom . The incidental music was very pleasant to listen to ; however I hated the songs , they were really uninspired , and nothing like the beautiful Tchaikovsky inspired melodies of the original . The characters were the strongest development here . Cinderella while still caring , had lost her sincerity , and a lot of her charm from the original , though she does wear some very pretty clothes . The Duke had some truly funny moments but they were n't enough to save the \textbf{cinema} , likewise with Prudence and the king . As I mentioned , the mice were the redeeming merits of the movie , as they alone contributed to the film 's cuteness . I have to say also the animation is colourful and above average , and the voice acting was surprisingly good . All in all , a cute , if unoriginal sequel , that was \textbf{tempered} by the songs and a lack of a story . 4/10 for the mice , the \textbf{voices} acting , the animation and some pretty dresses . Bethany Cox
\end{minipage} }
Correct label: negative. \\
Model confidence on original example: 90.3.
\end{figure*}

\begin{figure*}
Normally trained model, example 2 \\
\fbox{ \begin{minipage}{\textwidth}
Original: When I was younger , I liked this show , but now ... BLECCH ! ! ! This show is sappy , \textbf{badly} written , and rarely \textbf{funny} . The three leads \textbf{were} all good actors and \textbf{funny} men ( Saget 's stand up was a lot better than the stuff this show came up with , as was Coulier a better \textbf{stand} up , and Stamos was a better than average actor ) . After a while , Stamos wanted off the show because it wanted to do more serious stuff ( who could blame him ? ) . The show eventually got cancelled when many of the actors demanded more money . Here are a few things that drive me crazy about the show : 1 . The \textbf{catch} phrases- How many times can one person put up with tiring \textbf{catch} phrases like with 'how rude ' , 'you got it \textbf{dude} ' , 'nerdbomber ' , 'cut it out ' and 'have mercy ' in a 24 hour time period ? 2 . Kimmy Gibler- the most \textbf{annoying} character ever written for television . 3 . The writing- \textbf{stale} and cliched as an oreo \textbf{cookie} . There is good \textbf{cliched} writing and \textbf{bad} \textbf{cliched} writing . Full House had \textbf{bad} cliched writing . 4 . Three men living together in San Francisco- Enough said . 5 . Unrealistic stuff- Too much to recall . 6 . Trendy kids- The \textbf{girls} had all the \textbf{latest} mall fashions and you can see posters of trendy recording \textbf{artists} they would be into . Now this show is on Nick @ Nite . I would hardly call it a classic . I have nothing bad to say about the people involved since I think many of them are \textbf{talented} in their own right . But this show was just so sugary sweet , I could n't stand it after a while .
\end{minipage} }
\fbox{ \begin{minipage}{\textwidth}
Perturbed: When I was younger , I liked this show , but now ... BLECCH ! ! ! This show is sappy , \textbf{desperately} written , and rarely \textbf{hilarious} . The three leads \textbf{was} all good actors and \textbf{hilarious} men ( Saget 's stand up was a lot better than the stuff this show came up with , as was Coulier a better \textbf{stands} up , and Stamos was a better than average actor ) . After a while , Stamos wanted off the show because it wanted to do more serious stuff ( who could blame him ? ) . The show eventually got cancelled when many of the actors demanded more money . Here are a few things that drive me crazy about the show : 1 . The \textbf{captures} phrases- How many times can one person put up with tiring \textbf{captures} phrases like with 'how rude ' , 'you got it \textbf{bro} ' , 'nerdbomber ' , 'cut it out ' and 'have mercy ' in a 24 hour time period ? 2 . Kimmy Gibler- the most \textbf{exasperating} character ever written for television . 3 . The writing- \textbf{obsolete} and cliched as an oreo \textbf{cookies} . There is good \textbf{corny} writing and \textbf{wicked} \textbf{corny} writing . Full House had \textbf{wicked} cliched writing . 4 . Three men living together in San Francisco- Enough said . 5 . Unrealistic stuff- Too much to recall . 6 . Trendy kids- The \textbf{daughters} had all the \textbf{last} mall fashions and you can see posters of trendy recording \textbf{artistes} they would be into . Now this show is on Nick @ Nite . I would hardly call it a classic . I have nothing bad to say about the people involved since I think many of them are \textbf{genius} in their own right . But this show was just so sugary sweet , I could n't stand it after a while .
\end{minipage} }
Correct label: negative. \\
Model confidence on original example: 98.9.
\end{figure*}

\begin{figure*}
Normally trained model, example 3 \\
\fbox{ \begin{minipage}{\textwidth}
Original: Jim Carrey \textbf{shines} in this \textbf{beautiful} movie . This is \textbf{now} one of my \textbf{favorite} movies . I read all about the making and I \textbf{thought} it was \textbf{incredible} how the \textbf{did} it . I \textbf{ca} n't wait \textbf{till} this comes out on DVD . I \textbf{saw} this in \textbf{theaters} \textbf{so} many \textbf{times} , I \textbf{ca} n't even count how \textbf{times} I 've \textbf{seen} it .
\end{minipage} }
\fbox{ \begin{minipage}{\textwidth}
Perturbed: Jim Carrey \textbf{stars} in this \textbf{handsome} movie . This is \textbf{currently} one of my \textbf{favourite} movies . I read all about the making and I \textbf{figured} it was \textbf{unthinkable} how the \textbf{am} it . I \textbf{could} n't wait \textbf{unless} this comes out on DVD . I \textbf{watched} this in \textbf{theatres} \textbf{too} many \textbf{time} , I \textbf{could} n't even count how \textbf{period} I 've \textbf{watched} it .
\end{minipage} }
Correct label: positive. \\
Model confidence on original example: 100.0.
\end{figure*}

\begin{figure*}
Normally trained model, example 4 \\
\fbox{ \begin{minipage}{\textwidth}
Original: Did anyone read the script . This has to be some of the \textbf{worst} writing and directing of the entire year . Three great \textbf{actors} , Paul Giamatti , Rachel Weisz and Miranda Richardson could n't pull this one out . About two-thirds it looked like Giamatti eyes were \textbf{saying} , I ca n't believe I signed the contract . It 's not the \textbf{worst} movie I ever saw , but it 's on the \textbf{really} really \textbf{bad} Christmas movie list . Not enough lines , but what else can be said ? Okay , the movie just does n't move with Vaughn 's con-man dialogue , his character is just a \textbf{creepy} \textbf{guy} that you \textbf{just} ca n't get \textbf{past} . It was just a \textbf{lackluster} walk \textbf{through} , that no one seemed to be able to get into .
\end{minipage} }
\fbox{ \begin{minipage}{\textwidth}
Perturbed: Did anyone read the script . This has to be some of the \textbf{toughest} writing and directing of the entire year . Three great \textbf{protagonists} , Paul Giamatti , Rachel Weisz and Miranda Richardson could n't pull this one out . About two-thirds it looked like Giamatti eyes were \textbf{telling} , I ca n't believe I signed the contract . It 's not the \textbf{toughest} movie I ever saw , but it 's on the \textbf{truly} really \textbf{wicked} Christmas movie list . Not enough lines , but what else can be said ? Okay , the movie just does n't move with Vaughn 's con-man dialogue , his character is just a \textbf{terrifying} \textbf{buddy} that you \textbf{only} ca n't get \textbf{last} . It was just a \textbf{puny} walk \textbf{throughout} , that no one seemed to be able to get into .
\end{minipage} }
Correct label: negative. \\
Model confidence on original example: 100.0.
\end{figure*}

\begin{figure*}
Normally trained model, example 5 \\
\fbox{ \begin{minipage}{\textwidth}
Original: Yes I have rated this film as one star \textbf{awful} . Yet , it will be in my rotation of Christmas \textbf{movies} henceforth . This truly is so bad it 's good . This is another K.Gordon Murray \textbf{production} ( read : buys a really cheap/bad Mexican movie , spends \textbf{zero} money getting it dubbed into English and releases it at kiddie matinées in the mid 1960 's . ) It 's a shame I stumbled on this so late in life as I 'm sure some `` mood enhancers '' would make this an even better experience . I 'm not going to rehash what so many of the other reviewers have already said , a Christmas movie with Merlin , the Devil , mechanical wind-up reindeer and some of the most \textbf{pathetic} child actors I have ever seen bar none . I plan on running this over the holidays back to back with Kelsey Grammar 's `` A Christmas Carol '' . Truly a \textbf{holiday} experience made in Hell . Now if I can only find `` To All A Goodnight ( aka Slayride ) '' on DVD I 'll have a triple feature that ca n't be beat . You have to \textbf{see} this movie . It moves so slowly that I defy you not to touch the fast forward button-especially on the two dance routines ! This thing reeks like an expensive bleu cheese-guess you have to get past the \textbf{stink} to enjoy the experience . Feliz Navidad amigos !
\end{minipage} }
\fbox{ \begin{minipage}{\textwidth}
Perturbed: Yes I have rated this film as one star \textbf{horrifying} . Yet , it will be in my rotation of Christmas \textbf{cinema} henceforth . This truly is so bad it 's good . This is another K.Gordon Murray \textbf{producing} ( read : buys a really cheap/bad Mexican movie , spends \textbf{nought} money getting it dubbed into English and releases it at kiddie matinées in the mid 1960 's . ) It 's a shame I stumbled on this so late in life as I 'm sure some `` mood enhancers '' would make this an even better experience . I 'm not going to rehash what so many of the other reviewers have already said , a Christmas movie with Merlin , the Devil , mechanical wind-up reindeer and some of the most \textbf{lamentable} child actors I have ever seen bar none . I plan on running this over the holidays back to back with Kelsey Grammar 's `` A Christmas Carol '' . Truly a \textbf{festive} experience made in Hell . Now if I can only find `` To All A Goodnight ( aka Slayride ) '' on DVD I 'll have a triple feature that ca n't be beat . You have to \textbf{admire} this movie . It moves so slowly that I defy you not to touch the fast forward button-especially on the two dance routines ! This thing reeks like an expensive bleu cheese-guess you have to get past the \textbf{scent} to enjoy the experience . Feliz Navidad amigos !
\end{minipage} }
Correct label: negative. \\
Model confidence on original example: 97.0.
\end{figure*}

\begin{figure*}
Normally trained model, example 6 \\
\fbox{ \begin{minipage}{\textwidth}
Original: This show is quick-witted , \textbf{colorful} , \textbf{dark} yet fun , hip and \textbf{still} somehow clean . The cast , including an awesome rotation of special \textbf{guests} ( i.e . Molly Shannon , Paul Rubens , The-Stapler-Guy-From-Office-Space ) is \textbf{electric} . It 's got murder , romance , \textbf{family} , AND zombies without ever coming off as cartoony ... Somehow . You really connect with these characters . The whole production is an unlikely \textbf{magic} act that left me , something of a skeptic if I do say so myself , totally engrossed and \textbf{coming} back for more every Wednesday night . I \textbf{just} re-read this and it sounds a little like somebody paid me to write it . It really is that good . I just heard a \textbf{rumor} that it was being canceled so I \textbf{thought} I 'd send off a flare of good \textbf{will} . This is one of those shows that goes under the radar because the network suits ca n't figure out how to make it sexy and sell cars with it . Do yourself a huge favor , if you have n't already , and enjoy this gem while it lasts . OK so one more thing . This show is clever . What that means is that every armchair critic/ '' writer '' in Hollywood is gon na insert a stick up their youknowwhat before they sit down to watch it , defending themselves with an `` I could 've written that '' \textbf{type} speech to absolutely nobody in their lonely renovated Hollywood hotel room . In other words : the internet . This is a general interest/anonymous website . Before you give your Wednesday TV hour to Dirty Sexy Money or Next Hot Model reruns or whatever other out and out tripe these internet `` critics '' are n't commenting on , \textbf{give} my fave ' show a spin . It 's fun . Good , unpretentious fun .
\end{minipage} }
\fbox{ \begin{minipage}{\textwidth}
Perturbed: This show is quick-witted , \textbf{colored} , \textbf{darkened} yet fun , hip and \textbf{even} somehow clean . The cast , including an awesome rotation of special \textbf{guest} ( i.e . Molly Shannon , Paul Rubens , The-Stapler-Guy-From-Office-Space ) is \textbf{electricity} . It 's got murder , romance , \textbf{relatives} , AND zombies without ever coming off as cartoony ... Somehow . You really connect with these characters . The whole production is an unlikely \textbf{hallucinogenic} act that left me , something of a skeptic if I do say so myself , totally engrossed and \textbf{come} back for more every Wednesday night . I \textbf{merely} re-read this and it sounds a little like somebody paid me to write it . It really is that good . I just heard a \textbf{rumour} that it was being canceled so I \textbf{figured} I 'd send off a flare of good \textbf{willpower} . This is one of those shows that goes under the radar because the network suits ca n't figure out how to make it sexy and sell cars with it . Do yourself a huge favor , if you have n't already , and enjoy this gem while it lasts . OK so one more thing . This show is clever . What that means is that every armchair critic/ '' writer '' in Hollywood is gon na insert a stick up their youknowwhat before they sit down to watch it , defending themselves with an `` I could 've written that '' \textbf{typing} speech to absolutely nobody in their lonely renovated Hollywood hotel room . In other words : the internet . This is a general interest/anonymous website . Before you give your Wednesday TV hour to Dirty Sexy Money or Next Hot Model reruns or whatever other out and out tripe these internet `` critics '' are n't commenting on , \textbf{lend} my fave ' show a spin . It 's fun . Good , unpretentious fun .
\end{minipage} }
Correct label: positive. \\
Model confidence on original example: 93.7.
\end{figure*}

\begin{figure*}
Normally trained model, example 7 \\
\fbox{ \begin{minipage}{\textwidth}
Original: This was a hit in the South By Southwest ( SXSW ) Film festival in Austin last year , and features a fine \textbf{cast} headed up by E.R . 's Gloria Reuben , and a scenery-chewing John Glover . Though \textbf{shot} on a \textbf{small} budget in NYC , the film looks and \textbf{sounds} fabulous , and takes us on a behind the scenes whirl through the \textbf{rehearsal} and mounting of what \textbf{actors} call `` The Scottish Play , '' as a reference to the word `` Macbeth '' is \textbf{thought} to bring on the \textbf{play} 's \textbf{ancient} curse . The acting company exhibits all the \textbf{emotions} of the play \textbf{itself} , lust , jealousy , rage , suspicion , and a bit of fun as well . The \textbf{games} \textbf{begin} when an accomplished actor is replaced ( in the lead role ) by a well-known `` pretty face '' from the TV soap opera scene in order to draw bigger crowds . The green-eyed monster takes over from there , and the \textbf{drama} unfolds \textbf{nicely} . Fine soundtrack , and good performances all around . The DVD \textbf{includes} director 's \textbf{commentary} and some deleted scenes as \textbf{well} .
\end{minipage} }
\fbox{ \begin{minipage}{\textwidth}
Perturbed: This was a hit in the South By Southwest ( SXSW ) Film festival in Austin last year , and features a fine \textbf{casting} headed up by E.R . 's Gloria Reuben , and a scenery-chewing John Glover . Though \textbf{murdered} on a \textbf{tiny} budget in NYC , the film looks and \textbf{sound} fabulous , and takes us on a behind the scenes whirl through the \textbf{repeating} and mounting of what \textbf{actresses} call `` The Scottish Play , '' as a reference to the word `` Macbeth '' is \textbf{thinking} to bring on the \textbf{toy} 's \textbf{old} curse . The acting company exhibits all the \textbf{thrills} of the play \textbf{yourselves} , lust , jealousy , rage , suspicion , and a bit of fun as well . The \textbf{play} \textbf{starts} when an accomplished actor is replaced ( in the lead role ) by a well-known `` pretty face '' from the TV soap opera scene in order to draw bigger crowds . The green-eyed monster takes over from there , and the \textbf{theatre} unfolds \textbf{politely} . Fine soundtrack , and good performances all around . The DVD \textbf{contains} director 's \textbf{remark} and some deleted scenes as \textbf{good} .
\end{minipage} }
Correct label: positive. \\
Model confidence on original example: 97.8.
\end{figure*}

\begin{figure*}
Normally trained model, example 8 \\
\fbox{ \begin{minipage}{\textwidth}
Original: As a young black/latina woman I am always \textbf{searching} for movies that represent the experiences and lives of people like me . Of course when I saw this movie at the video store I thought I would enjoy it ; unfortunately , I did n't . Although the \textbf{topics} presented in the film are interesting and relevant , the story was simply not properly developed . The movie just kept dragging on and on and many of the characters that appear on screen just come and go without much to contribute to the overall film . Had the director done a better job interconnecting the scenes , perhaps I would have enjoyed it a bit more . Honestly , I would recommend a film like `` Raising Victor '' \textbf{over} this one any day . I just was not too impressed .
\end{minipage} }
\fbox{ \begin{minipage}{\textwidth}
Perturbed: As a young black/latina woman I am always \textbf{browsing} for movies that represent the experiences and lives of people like me . Of course when I saw this movie at the video store I thought I would enjoy it ; unfortunately , I did n't . Although the \textbf{themes} presented in the film are interesting and relevant , the story was simply not properly developed . The movie just kept dragging on and on and many of the characters that appear on screen just come and go without much to contribute to the overall film . Had the director done a better job interconnecting the scenes , perhaps I would have enjoyed it a bit more . Honestly , I would recommend a film like `` Raising Victor '' \textbf{finished} this one any day . I just was not too impressed .
\end{minipage} }
Correct label: negative. \\
Model confidence on original example: 65.4.
\end{figure*}

\begin{figure*}
Normally trained model, example 9 \\
\fbox{ \begin{minipage}{\textwidth}
Original: If Fassbinder has made a worse film , I sure do n't want to see it ! Anyone who complains that his films are too talky and claustrophobic should be forced to view this , to learn to appreciate the more spare style he opted for in excellent films like `` The Bitter Tears Of Petra von Kant '' . This film bogs down with so much arty , quasi-symbolic images it looks like a parody of an `` art-film '' . The scene in the slaughterhouse and the scene where Elvira 's prostitute friend channel-surfs for what seems like ten minutes are just two of the most glaring examples of what makes this film a real test of the viewer 's endurance . But what really angers me about it are the few scenes which feature just Elvira and her ex-wife and/or her daughter . These are the only moments that display any real human emotion , and prove that at the core of this \textbf{horrible} film , there was an excellent film struggling to free itself . What a waste .
\end{minipage} }
\fbox{ \begin{minipage}{\textwidth}
Perturbed: If Fassbinder has made a worse film , I sure do n't want to see it ! Anyone who complains that his films are too talky and claustrophobic should be forced to view this , to learn to appreciate the more spare style he opted for in excellent films like `` The Bitter Tears Of Petra von Kant '' . This film bogs down with so much arty , quasi-symbolic images it looks like a parody of an `` art-film '' . The scene in the slaughterhouse and the scene where Elvira 's prostitute friend channel-surfs for what seems like ten minutes are just two of the most glaring examples of what makes this film a real test of the viewer 's endurance . But what really angers me about it are the few scenes which feature just Elvira and her ex-wife and/or her daughter . These are the only moments that display any real human emotion , and prove that at the core of this \textbf{gruesome} film , there was an excellent film struggling to free itself . What a waste .
\end{minipage} }
Correct label: negative. \\
Model confidence on original example: 65.8.
\end{figure*}

\begin{figure*}
Normally trained model, example 10 \\
\fbox{ \begin{minipage}{\textwidth}
Original: This film is the \textbf{worst} excuse for a motion picture I have EVER seen . To begin , I 'd \textbf{like} to say the the front cover of this film is by all means misleading , if you think you are about to see a truly \textbf{scary} horror film with a monster clown , you are soooo wrong . In fact the killers face does n't even slightly resemble the front cover , it 's just an image they must have found on Google and thought it looked \textbf{cool} . Speaking of things they \textbf{found} and thought it looked cool , there is a scene in this film where some of the gang are searching for the friend in the old woods , then suddenly the screen chops to a scene where there is a mother deer nurturing it 's young in a glisten of sunlight ... I mean \textbf{seriously} WTF ? ? ? How is this relevant to the dark woods they are wandering through ? I bought this film from a man at a market hoping it would be entertaining , if it was n't horror then at least it would be funny right ? WRONG ! The next day I GAVE it to my work colleague \textbf{ridding} myself from the plague named S.I.C.K Bottom line is : Do n't SEE THIS FILM ! ! !
\end{minipage} }
\fbox{ \begin{minipage}{\textwidth}
Perturbed: This film is the \textbf{toughest} excuse for a motion picture I have EVER seen . To begin , I 'd \textbf{love} to say the the front cover of this film is by all means misleading , if you think you are about to see a truly \textbf{terrifying} horror film with a monster clown , you are soooo wrong . In fact the killers face does n't even slightly resemble the front cover , it 's just an image they must have found on Google and thought it looked \textbf{groovy} . Speaking of things they \textbf{discovered} and thought it looked cool , there is a scene in this film where some of the gang are searching for the friend in the old woods , then suddenly the screen chops to a scene where there is a mother deer nurturing it 's young in a glisten of sunlight ... I mean \textbf{deeply} WTF ? ? ? How is this relevant to the dark woods they are wandering through ? I bought this film from a man at a market hoping it would be entertaining , if it was n't horror then at least it would be funny right ? WRONG ! The next day I GAVE it to my work colleague \textbf{liberating} myself from the plague named S.I.C.K Bottom line is : Do n't SEE THIS FILM ! ! !
\end{minipage} }
Correct label: negative. \\
Model confidence on original example: 95.7.
\end{figure*}

\begin{figure*}
Certifiably robust model, example 1 \\
\fbox{ \begin{minipage}{\textwidth}
Original: Rohmer returns to his historical dramas in the real story of Grace Elliot , an Englishwoman who stayed in France during the apex of the French Revolution . One always suspected that Rohmer was a conservative , but who knew he was such a red-blooded reactionary . If you can put aside Rohmer 's unabashed defense of the monarchy ( and that is not an easy thing to do , given that , for instance , the French lower classes are portrayed here as \textbf{hideous} louts ) , this is actually an elegant , intelligent and polished movie . Lacking the money for a big cinematic recreation of 18th century France , Rohmer has instead the actors play against obvious painted cardboards . It is a blatantly artificial conceit , but it somehow works . And newcomer Lucy Russell succeeds in making sympathetic a character that should n't be .
\end{minipage} }
\fbox{ \begin{minipage}{\textwidth}
Perturbed: Rohmer returns to his historical dramas in the real story of Grace Elliot , an Englishwoman who stayed in France during the apex of the French Revolution . One always suspected that Rohmer was a conservative , but who knew he was such a red-blooded reactionary . If you can put aside Rohmer 's unabashed defense of the monarchy ( and that is not an easy thing to do , given that , for instance , the French lower classes are portrayed here as \textbf{ghastly} louts ) , this is actually an elegant , intelligent and polished movie . Lacking the money for a big cinematic recreation of 18th century France , Rohmer has instead the actors play against obvious painted cardboards . It is a blatantly artificial conceit , but it somehow works . And newcomer Lucy Russell succeeds in making sympathetic a character that should n't be .
\end{minipage} }
Correct label: positive. \\
Model confidence on original example: 65.2.
\end{figure*}

\begin{figure*}
Certifiably robust model, example 2 \\
\fbox{ \begin{minipage}{\textwidth}
Original: Warning Spoiler . . . I have to agree with you , it was almost there . This was such a bad movie , about such and interesting true story . It had such promise , but the acting was ridiculous at best . Some sets were \textbf{beautiful} and realistic . Others are something out of a theme park . I found myself laughing as I watched , what was suppose to be , serious scenes . I really wanted to like this movie , but I could n't . The best part was the fight between friends that ended with the `` King '' dying . I liked the Queens ' punishment . And , the final shot made a beautiful picture , though . There are so many better movies to watch . I do n't recommend this .
\end{minipage} }
\fbox{ \begin{minipage}{\textwidth}
Perturbed: Warning Spoiler . . . I have to agree with you , it was almost there . This was such a bad movie , about such and interesting true story . It had such promise , but the acting was ridiculous at best . Some sets were \textbf{marvelous} and realistic . Others are something out of a theme park . I found myself laughing as I watched , what was suppose to be , serious scenes . I really wanted to like this movie , but I could n't . The best part was the fight between friends that ended with the `` King '' dying . I liked the Queens ' punishment . And , the final shot made a beautiful picture , though . There are so many better movies to watch . I do n't recommend this .
\end{minipage} }
Correct label: negative. \\
Model confidence on original example: 56.7.
\end{figure*}

\begin{figure*}
Certifiably robust model, example 3 \\
\fbox{ \begin{minipage}{\textwidth}
Original: `` Raw Force '' is like an ultra-sleazy and perverted version of Love Boat , with additional Kung Fu fights , demented cannibalistic monks , white slaves trade , energetic zombies and a whole lot of \textbf{lousy} acting performances . No wonder this movie was included in the recently released `` Grindhouse Experience 20 movie box-set '' . It 's got everything exploitation fanatics are looking for , blend in a totally \textbf{incoherent} and seemingly improvised \textbf{script} ! The production values are extremely poor and the technical aspects are \textbf{pathetic} , but the amounts of gratuitous violence \& sex can hardly be described . The film opens at a tropically sunny location called Warriors Island , where a troop of sneering monks raise the dead for no apparent reason other than to turn them into Kung Fu fighters . The monks also buy sexy slaves from a sleazy Hitler look-alike businessman , supposedly because the women 's flesh supplies them with the required powers to increase their zombie army . Tourists on a passing cruise ship , among them three martial arts fighters , a female LA cop and a whole bunch of \textbf{ravishing} but dim-witted ladies , are attacked by the Hitler guy 's goons because they were planning an excursion to Warriors Island . Their lifeboat washes ashore the island anyway , and the \textbf{monks} challenge the survivors to a fighting test with their zombies . Okay , how does that sound for a crazy midnight horror movie mess ? It 's not over yet , because `` Raw Force '' also has piranhas , wild boat orgies , Cameron Mitchell in yet another embarrassing lead role and 70 's exploitation duchess Camille Keaton ( `` I spit on your Grave '' ) in an utterly insignificant cameo appearance . There 's loads of \textbf{badly} realized gore , including \textbf{axe} massacres and decapitations , hammy jokes and bad taste romance . The trash-value of this movie will literally leave you speechless . The evil monks ' background remains , naturally , unexplained and they do n't \textbf{even} become punished for their questionable hobbies . Maybe that 's why the movie stops with `` To Be Continued '' , instead of with `` The End '' . The sequel never came , unless it 's so obscure IMDb does n't even list it .
\end{minipage} }
\fbox{ \begin{minipage}{\textwidth}
Perturbed: `` Raw Force '' is like an ultra-sleazy and perverted version of Love Boat , with additional Kung Fu fights , demented cannibalistic monks , white slaves trade , energetic zombies and a whole lot of \textbf{miserable} acting performances . No wonder this movie was included in the recently released `` Grindhouse Experience 20 movie box-set '' . It 's got everything exploitation fanatics are looking for , blend in a totally \textbf{inconsistent} and seemingly improvised \textbf{scenario} ! The production values are extremely poor and the technical aspects are \textbf{pitiable} , but the amounts of gratuitous violence \& sex can hardly be described . The film opens at a tropically sunny location called Warriors Island , where a troop of sneering monks raise the dead for no apparent reason other than to turn them into Kung Fu fighters . The monks also buy sexy slaves from a sleazy Hitler look-alike businessman , supposedly because the women 's flesh supplies them with the required powers to increase their zombie army . Tourists on a passing cruise ship , among them three martial arts fighters , a female LA cop and a whole bunch of \textbf{gorgeous} but dim-witted ladies , are attacked by the Hitler guy 's goons because they were planning an excursion to Warriors Island . Their lifeboat washes ashore the island anyway , and the \textbf{monk} challenge the survivors to a fighting test with their zombies . Okay , how does that sound for a crazy midnight horror movie mess ? It 's not over yet , because `` Raw Force '' also has piranhas , wild boat orgies , Cameron Mitchell in yet another embarrassing lead role and 70 's exploitation duchess Camille Keaton ( `` I spit on your Grave '' ) in an utterly insignificant cameo appearance . There 's loads of \textbf{sorely} realized gore , including \textbf{ax} massacres and decapitations , hammy jokes and bad taste romance . The trash-value of this movie will literally leave you speechless . The evil monks ' background remains , naturally , unexplained and they do n't \textbf{also} become punished for their questionable hobbies . Maybe that 's why the movie stops with `` To Be Continued '' , instead of with `` The End '' . The sequel never came , unless it 's so obscure IMDb does n't even list it .
\end{minipage} }
Correct label: negative. \\
Model confidence on original example: 72.5.
\end{figure*}

\begin{figure*}
Certifiably robust model, example 4 \\
\fbox{ \begin{minipage}{\textwidth}
\small
Original: It was the Sixties , and anyone with long hair and a hip , distant attitude could get money to make a movie . That 's how Michael Sarne , director of this colossal \textbf{flop} , was able to get the job . Sarne is one of the most supremely untalented people ever given a dollar to make a movie . In fact , the whole studio must have been tricked into agreeing to hire a guy who had made exactly one previous film , a terribly precious 60's-hip black and white featurette called Joanna . That film starred the similarly talentless actress/waif Genevieve Waite who could barely speak an entire line without breaking into some inappropriate facial expression or bat-like twitter . Sarne , who was probably incapable of directing a cartoon , never mind a big-budget Hollywood film , was in way over his head . David Giler 's book is the best place to go to find out how the faux-infant terrible Sarne was able to pull the wool over everyone 's eyes . If there is ever an historical marker which indicates the superficiality and shallowness of an era , Myra Breckinridge provides that marker . It embodies the emptiness and mindless excess of a decade which is more often remembered for a great sea-change in the body politic . Breckinridge is a touchstone of another , equally important vein . Watch this movie and you 'll get a different perspective on the less-often mentioned vacuity of spirit which so often passed for talent during those years . Many reviewers have spoken about the inter-cutting of footage from other films , especially older ones . Some actually liked these clunky `` comments '' on what was taking place in the movie , others found them \textbf{senseless} , \textbf{annoying} , and obtrusive , though since the film is so bad itself any intrusion would have to be an improvement . In my opinion , the real reason Michael Sarne put so many film clips into Myra Brekinridge was to \textbf{paper} over the bottomless insufficiency of wit and imagination that he possessed . That is to say , Sarne was so imagination-challenged that he just threw these clips in to fill space and take up time . They were n't inspiration , they were desperation . His writing skills were nonexistent , and David Giler had wisely stepped away from the project as one might from a ticking bomb , so Sarne was left to actually try and make a movie , and he could n't . It was beyond his slim capabilities . Hence the introduction of what seems like one half of an entire film 's worth of clips . The ghosts of writers and directors - many long since passed on - were called upon to fix this calamitous flopperoo because Sarne sure as heck was n't able to . This was what he came up with on those days he sat on the set and thought for eight hours while the entire cast and crew ( not to mention the producers and the \textbf{accountants} ) cooled their heels and waited for something , some \textbf{great} spark of imagination , a hint of originality , a soupcon of wit , to crackle forth from the brow of Zeus . Um , oops . No Zeus + no imagination + no sparks = millions of little dollar bills with tiny wings - each made from the hundreds of licensing agreements required to use the clips - flying out the window . Bye-bye . As for myself , I hated the film clips . They denigrated Sarne 's many betters , poked fun at people whose talents - even those whose skills were not \textbf{great} - far outstripped the abilities of the director and so ultimately served to show how lacking he was in inspiration , originality - and even of plain competency - compared to even the cheesiest of them .
\end{minipage} }
\fbox{ \begin{minipage}{\textwidth}
\small
Perturbed: It was the Sixties , and anyone with long hair and a hip , distant attitude could get money to make a movie . That 's how Michael Sarne , director of this colossal \textbf{bankruptcy} , was able to get the job . Sarne is one of the most supremely untalented people ever given a dollar to make a movie . In fact , the whole studio must have been tricked into agreeing to hire a guy who had made exactly one previous film , a terribly precious 60's-hip black and white featurette called Joanna . That film starred the similarly talentless actress/waif Genevieve Waite who could barely speak an entire line without breaking into some inappropriate facial expression or bat-like twitter . Sarne , who was probably incapable of directing a cartoon , never mind a big-budget Hollywood film , was in way over his head . David Giler 's book is the best place to go to find out how the faux-infant terrible Sarne was able to pull the wool over everyone 's eyes . If there is ever an historical marker which indicates the superficiality and shallowness of an era , Myra Breckinridge provides that marker . It embodies the emptiness and mindless excess of a decade which is more often remembered for a great sea-change in the body politic . Breckinridge is a touchstone of another , equally important vein . Watch this movie and you 'll get a different perspective on the less-often mentioned vacuity of spirit which so often passed for talent during those years . Many reviewers have spoken about the inter-cutting of footage from other films , especially older ones . Some actually liked these clunky `` comments '' on what was taking place in the movie , others found them \textbf{wanton} , \textbf{troublesome} , and obtrusive , though since the film is so bad itself any intrusion would have to be an improvement . In my opinion , the real reason Michael Sarne put so many film clips into Myra Brekinridge was to \textbf{papers} over the bottomless insufficiency of wit and imagination that he possessed . That is to say , Sarne was so imagination-challenged that he just threw these clips in to fill space and take up time . They were n't inspiration , they were desperation . His writing skills were nonexistent , and David Giler had wisely stepped away from the project as one might from a ticking bomb , so Sarne was left to actually try and make a movie , and he could n't . It was beyond his slim capabilities . Hence the introduction of what seems like one half of an entire film 's worth of clips . The ghosts of writers and directors - many long since passed on - were called upon to fix this calamitous flopperoo because Sarne sure as heck was n't able to . This was what he came up with on those days he sat on the set and thought for eight hours while the entire cast and crew ( not to mention the producers and the \textbf{accounting} ) cooled their heels and waited for something , some \textbf{magnificent} spark of imagination , a hint of originality , a soupcon of wit , to crackle forth from the brow of Zeus . Um , oops . No Zeus + no imagination + no sparks = millions of little dollar bills with tiny wings - each made from the hundreds of licensing agreements required to use the clips - flying out the window . Bye-bye . As for myself , I hated the film clips . They denigrated Sarne 's many betters , poked fun at people whose talents - even those whose skills were not \textbf{magnificent} - far outstripped the abilities of the director and so ultimately served to show how lacking he was in inspiration , originality - and even of plain competency - compared to even the cheesiest of them .
\end{minipage} }
Correct label: negative. \\
Model confidence on original example: 55.9.
\end{figure*}

\begin{figure*}
Certifiably robust model, example 5 \\
\fbox{ \begin{minipage}{\textwidth}
Original: I totally got drawn into this and could n't wait for each episode . The \textbf{acting} brought to life how emotional a missing person in the family must be , together with the effects it would have on those closest . The only problem we as a family had was how quickly it was all 'explained ' at the end . We could n't hear clearly what was said and have no idea what Gary 's part in the whole thing was ? Why did Kyle phone him and why did he go along with it ? Having invested in a series for five hours we felt \textbf{cheated} that only five minutes was kept back for the conclusion . I have asked around and none of my friends who watched it were any the wiser either . Very strange but maybe we missed something crucial ? ? ? ?
\end{minipage} }
\fbox{ \begin{minipage}{\textwidth}
Perturbed: I totally got drawn into this and could n't wait for each episode . The \textbf{behaving} brought to life how emotional a missing person in the family must be , together with the effects it would have on those closest . The only problem we as a family had was how quickly it was all 'explained ' at the end . We could n't hear clearly what was said and have no idea what Gary 's part in the whole thing was ? Why did Kyle phone him and why did he go along with it ? Having invested in a series for five hours we felt \textbf{hoodwinked} that only five minutes was kept back for the conclusion . I have asked around and none of my friends who watched it were any the wiser either . Very strange but maybe we missed something crucial ? ? ? ?
\end{minipage} }
Correct label: positive. \\
Model confidence on original example: 50.3.
\end{figure*}

\begin{figure*}
Certifiably robust model, example 6 \\
\fbox{ \begin{minipage}{\textwidth}
Original: One of the more sensible comedies to hit the Hindi film screens . A remake of Priyadarshans 80s Malayalam hit Boeing Boeing , which in turn was a remake of the 60s Hollywoon hit of the same name , Garam Masala elevates the standard of comedies in Hindi Cinema . Akshay Kumar has once again proved his is one of the best super stars of Hindi cinema who can do comedy . He has combined well with the new hunk John Abraham . However John still remains in Akshays shadows and \textbf{fails} to rise to the occasion . The new gals are cute and do complete justice to their roles . A must watch comedy . Leave your brains away and \textbf{laugh} for 2 hrs ! ! ! ! After all laughter is the best medicine ! Ask Priyadarshan and Akshay Kumar ! ! ! ! !
\end{minipage} }
\fbox{ \begin{minipage}{\textwidth}
Perturbed: One of the more sensible comedies to hit the Hindi film screens . A remake of Priyadarshans 80s Malayalam hit Boeing Boeing , which in turn was a remake of the 60s Hollywoon hit of the same name , Garam Masala elevates the standard of comedies in Hindi Cinema . Akshay Kumar has once again proved his is one of the best super stars of Hindi cinema who can do comedy . He has combined well with the new hunk John Abraham . However John still remains in Akshays shadows and \textbf{neglects} to rise to the occasion . The new gals are cute and do complete justice to their roles . A must watch comedy . Leave your brains away and \textbf{laughed} for 2 hrs ! ! ! ! After all laughter is the best medicine ! Ask Priyadarshan and Akshay Kumar ! ! ! ! !
\end{minipage} }
Correct label: positive. \\
Model confidence on original example: 51.3.
\end{figure*}

\begin{figure*}
Certifiably robust model, example 7 \\
\fbox{ \begin{minipage}{\textwidth}
Original: DER TODESKING is not one of my \textbf{favorite} Jorg Buttgereit film - but still is an interesting film dealing with suicide and it 's reasons and ramifications . Those looking for a gore-fest , or exploitation in the style of the NEKROMANTIK films or SCHRAMM will probably be \textbf{disappointed} . DER TODESKING is definitely an `` art-house '' style film , so those that need linear , explainable narratives need not apply ... The basic concept of DER TODESKING is that there is an `` episode '' for each day of the week that revolves around a strange chain letter that apparently causes people to commit suicide , interspersed with scenes of a slowly decomposing corpse ... There are some very well done and thought provoking scenes , including the man talking about the `` problems '' with his wife , and the concert massacre ( which unfortunately lost some of it 's `` power '' on me , because I was too busy laughing at the SCORPIONS look-alike band on stage ... ) . But seriously - this is a sometimes \textbf{beautiful} ( the scene that shows different angles of that huge bridge is particularly effective - especially if you understand the significance of the scene , and that the names shown are of people that actually committed suicide from jumping from the bridge ... ) , sometimes confusing , sometimes silly ( the SHE WOLF OF THE SS rip-off is pretty amusing ) , sometimes \textbf{harrowing} ( I found the scene of the guy talking to the girl in the park about his wife particularly effective ) film that is more of an `` experience '' then just entertainment , as many of these `` art '' films are meant to be . Still , I did n't find DER TODESKING to be as strong as NEKROMANTIK or SCHRAMM , and would probably put it on relatively even footing with NEKROMANTIK 2 in terms of my personally `` enjoyment level '' . Definitely worth a look to any Buttgereit or `` art '' film fan . If you dig this type of film - check out SUBCONSCIOUS CRUELTY - in my opinion the BEST art-house/horror film that I 've seen . 7/10 for DER TODESKING
\end{minipage} }
\fbox{ \begin{minipage}{\textwidth}
Perturbed: DER TODESKING is not one of my \textbf{preferred} Jorg Buttgereit film - but still is an interesting film dealing with suicide and it 's reasons and ramifications . Those looking for a gore-fest , or exploitation in the style of the NEKROMANTIK films or SCHRAMM will probably be \textbf{disappointing} . DER TODESKING is definitely an `` art-house '' style film , so those that need linear , explainable narratives need not apply ... The basic concept of DER TODESKING is that there is an `` episode '' for each day of the week that revolves around a strange chain letter that apparently causes people to commit suicide , interspersed with scenes of a slowly decomposing corpse ... There are some very well done and thought provoking scenes , including the man talking about the `` problems '' with his wife , and the concert massacre ( which unfortunately lost some of it 's `` power '' on me , because I was too busy laughing at the SCORPIONS look-alike band on stage ... ) . But seriously - this is a sometimes \textbf{handsome} ( the scene that shows different angles of that huge bridge is particularly effective - especially if you understand the significance of the scene , and that the names shown are of people that actually committed suicide from jumping from the bridge ... ) , sometimes confusing , sometimes silly ( the SHE WOLF OF THE SS rip-off is pretty amusing ) , sometimes \textbf{dreadful} ( I found the scene of the guy talking to the girl in the park about his wife particularly effective ) film that is more of an `` experience '' then just entertainment , as many of these `` art '' films are meant to be . Still , I did n't find DER TODESKING to be as strong as NEKROMANTIK or SCHRAMM , and would probably put it on relatively even footing with NEKROMANTIK 2 in terms of my personally `` enjoyment level '' . Definitely worth a look to any Buttgereit or `` art '' film fan . If you dig this type of film - check out SUBCONSCIOUS CRUELTY - in my opinion the BEST art-house/horror film that I 've seen . 7/10 for DER TODESKING
\end{minipage} }
Correct label: positive. \\
Model confidence on original example: 61.5.
\end{figure*}

\begin{figure*}
Certifiably robust model, example 8 \\
\fbox{ \begin{minipage}{\textwidth}
Original: Growing up , Joe Strummer was a hero of mine , but even I was left cold by this film . For better and worse , The Future Is Unwritten is not a straightforward `` Behind the Music '' style documentary . Rather it is a biographical art film , chock full of interviews , performance footage , home movies , and mostly \textbf{pointless} animation sketches lifted from `` Animal Farm . '' The movie is coherent but overlong by about a half hour . The campfire format , while touching in thought , is actually pretty annoying in execution . First off , without titles , its hard to even know who half of these interviewees are . Secondly , who really needs to hear people like Bono , Johnny Depp , and John Cusack mouth butt licking hosannas about the man ? They were not relevant to Strummer 's life and their opinions add nothing to his story . This picture is at it 's best when Strummer , through taped interviews and conversation , touches on facets of his life most people did not know about : the suicide of his older brother , coming to terms with the death of his parents , the joy of fatherhood . To me , these were most moving because it showed Joe Strummer not as the punk icon we all knew and loved , but as a regular human being who had to deal with the joys and sorrows of life we all must face . There have been better , more straightforward documentaries about Strummer and The Clash . ( Westway , VH1 Legends , and Kurt Loder 's narrated MTV Documentary from the early 90 's come to mind . ) Joe Strummer : The Future Is Unwritten is for diehards only .
\end{minipage} }
\fbox{ \begin{minipage}{\textwidth}
Perturbed: Growing up , Joe Strummer was a hero of mine , but even I was left cold by this film . For better and worse , The Future Is Unwritten is not a straightforward `` Behind the Music '' style documentary . Rather it is a biographical art film , chock full of interviews , performance footage , home movies , and mostly \textbf{unnecessary} animation sketches lifted from `` Animal Farm . '' The movie is coherent but overlong by about a half hour . The campfire format , while touching in thought , is actually pretty annoying in execution . First off , without titles , its hard to even know who half of these interviewees are . Secondly , who really needs to hear people like Bono , Johnny Depp , and John Cusack mouth butt licking hosannas about the man ? They were not relevant to Strummer 's life and their opinions add nothing to his story . This picture is at it 's best when Strummer , through taped interviews and conversation , touches on facets of his life most people did not know about : the suicide of his older brother , coming to terms with the death of his parents , the joy of fatherhood . To me , these were most moving because it showed Joe Strummer not as the punk icon we all knew and loved , but as a regular human being who had to deal with the joys and sorrows of life we all must face . There have been better , more straightforward documentaries about Strummer and The Clash . ( Westway , VH1 Legends , and Kurt Loder 's narrated MTV Documentary from the early 90 's come to mind . ) Joe Strummer : The Future Is Unwritten is for diehards only .
\end{minipage} }
Correct label: negative. \\
Model confidence on original example: 50.1.
\end{figure*}

\begin{figure*}
Certifiably robust model, example 9 \\
\fbox{ \begin{minipage}{\textwidth}
Original: The film begins with people on Earth discovering that their rocket to Mars had not been lost but was just drifting out in Space near out planet . When it 's retrieved , one of the crew members is ill , one is alive and the other two are missing . What happened to them is told through a flashback by the surviving member . While on Mars , the crew was apparently attacked by a whole host of very silly bug-eyed monsters . Oddly , while the sets were pretty good , the monsters were among the \textbf{silliest} I have seen on film . Plus , in an odd attempt at realism , the production used a process called `` Cinemagic '' . Unfortunately , this wonderful innovation just made the film look pretty cheap when they were on the surface of Mars AND the intensity of the redness practically made my eyes bleed -- it was THAT bad ! ! Despite all the cheese , the film did have a somewhat interesting plot as well as a good message about space travel . For lovers of the genre , it 's well worth seeing . For others , you may just find the whole thing rather silly -- see for yourself and decide . While by today 's standards this is n't an especially good sci-fi film , compared with the films being made at the time , it stacks up pretty well . PS -- When you watch the film , pay careful attention to Dr. Tremayne . He looks like the spitting image of Dr. Quest from the `` Jonny Quest '' cartoon ! Plus , he sounds and acts a lot like him , too .
\end{minipage} }
\fbox{ \begin{minipage}{\textwidth}
Perturbed: The film begins with people on Earth discovering that their rocket to Mars had not been lost but was just drifting out in Space near out planet . When it 's retrieved , one of the crew members is ill , one is alive and the other two are missing . What happened to them is told through a flashback by the surviving member . While on Mars , the crew was apparently attacked by a whole host of very silly bug-eyed monsters . Oddly , while the sets were pretty good , the monsters were among the \textbf{weirdest} I have seen on film . Plus , in an odd attempt at realism , the production used a process called `` Cinemagic '' . Unfortunately , this wonderful innovation just made the film look pretty cheap when they were on the surface of Mars AND the intensity of the redness practically made my eyes bleed -- it was THAT bad ! ! Despite all the cheese , the film did have a somewhat interesting plot as well as a good message about space travel . For lovers of the genre , it 's well worth seeing . For others , you may just find the whole thing rather silly -- see for yourself and decide . While by today 's standards this is n't an especially good sci-fi film , compared with the films being made at the time , it stacks up pretty well . PS -- When you watch the film , pay careful attention to Dr. Tremayne . He looks like the spitting image of Dr. Quest from the `` Jonny Quest '' cartoon ! Plus , he sounds and acts a lot like him , too .
\end{minipage} }
Correct label: negative. \\
Model confidence on original example: 54.8.
\end{figure*}

\begin{figure*}
Certifiably robust model, example 10 \\
\fbox{ \begin{minipage}{\textwidth}
Original: `` Sir '' John Gielgud must have become senile to star in a mess of a movie like this one . ; This is one of those films , I suppose , that is considered `` art , '' but do n't be fooled ... ..it 's \textbf{garbage} . Stick to the `` art '' you can admire in a frame because the films that are labeled as such are usually unintelligible forgeries like this . In this masterpiece , Giegud recites Shakespeare 's `` The Tempest '' while the camera pans away to nude people . one of them a little kid \textbf{urinating} in a swimming pool . Wow , this is heady stuff and real `` art , '' ai n't it ? ? That 's just one example . Most of the story makes no sense , is impossible to follow and , hence , is one that Liberal critics are afraid to say they did n't `` understand '' so they give it high marks to save their phony egos . You want Shakespeare ? Read his books .
\end{minipage} }
\fbox{ \begin{minipage}{\textwidth}
Perturbed: `` Sir '' John Gielgud must have become senile to star in a mess of a movie like this one . ; This is one of those films , I suppose , that is considered `` art , '' but do n't be fooled ... ..it 's \textbf{refuse} . Stick to the `` art '' you can admire in a frame because the films that are labeled as such are usually unintelligible forgeries like this . In this masterpiece , Giegud recites Shakespeare 's `` The Tempest '' while the camera pans away to nude people . one of them a little kid \textbf{urinate} in a swimming pool . Wow , this is heady stuff and real `` art , '' ai n't it ? ? That 's just one example . Most of the story makes no sense , is impossible to follow and , hence , is one that Liberal critics are afraid to say they did n't `` understand '' so they give it high marks to save their phony egos . You want Shakespeare ? Read his books .
\end{minipage} }
Correct label: negative. \\
Model confidence on original example: 67.7.
\end{figure*}

\begin{figure*}
Data augmentation model, example 1 \\
\fbox{ \begin{minipage}{\textwidth}
Original: I hope whoever coached these losers on their accents was fired . The only high points are a few of the supporting characters , 3 of 5 of my favourites \textbf{were} killed off by the end of the season ( and one of them was a cat , to put that into perspective ) . The whole storyline is centered around sex , and nothing else . Sex with vampires , gay sex with \textbf{gay} vampires , gay sex with straight vampires , sex to score vampire blood , sex after drinking vampire blood , sex in front of vampires , vampire sex , non-vampire sex , sex because we 're scared of vampires , sex because we 're mad at vampires , sex because we just became a vampire , etc . Nothing against sex , it would just be nice if it were a little more subtle with being peppered into the storyline . Perhaps HAVE a storyline and then shoehorn some sex into it . But they did n't even bother to do that ... and Anna Paquin is a dizzy gap-tooth bitch . Either she sucks or her character \textbf{sucks} , I ca n't figure out which . Another part of the storyline that I find highly \textbf{implausible} is why 150 year old vampire Bill who seems to have his things together would be interested in someone like Sookie . She 's constantly flying off the handle at him for things he ca n't control . He leaves for two days and she already decides that he 's `` not coming back '' and suddenly has feelings for dog-man ? Give me a break . She 's supposed to be a 25 year old woman , not a 14 year old girl . People close to her are dying all over , and she 's got the brightest smile on her face because she just gave away her V-card to some dude because she ca n't read his mind ? As the main character of the story , I would 've hoped the show would do a little more to make her understandable and someone to invest your interest in , not someone you keep secretly hoping gets killed off or put into a coma . I ca n't find anything about her character that I like and even the fact that she can read minds is impressively \textbf{uninspiring} and not the least bit interesting . I will not be wasting my time with watching Season 2 come June .
\end{minipage} }
\fbox{ \begin{minipage}{\textwidth}
Perturbed: I hope whoever coached these losers on their accents was fired . The only high points are a few of the supporting characters , 3 of 5 of my favourites \textbf{was} killed off by the end of the season ( and one of them was a cat , to put that into perspective ) . The whole storyline is centered around sex , and nothing else . Sex with vampires , gay sex with \textbf{homosexual} vampires , gay sex with straight vampires , sex to score vampire blood , sex after drinking vampire blood , sex in front of vampires , vampire sex , non-vampire sex , sex because we 're scared of vampires , sex because we 're mad at vampires , sex because we just became a vampire , etc . Nothing against sex , it would just be nice if it were a little more subtle with being peppered into the storyline . Perhaps HAVE a storyline and then shoehorn some sex into it . But they did n't even bother to do that ... and Anna Paquin is a dizzy gap-tooth bitch . Either she sucks or her character \textbf{fears} , I ca n't figure out which . Another part of the storyline that I find highly \textbf{improbable} is why 150 year old vampire Bill who seems to have his things together would be interested in someone like Sookie . She 's constantly flying off the handle at him for things he ca n't control . He leaves for two days and she already decides that he 's `` not coming back '' and suddenly has feelings for dog-man ? Give me a break . She 's supposed to be a 25 year old woman , not a 14 year old girl . People close to her are dying all over , and she 's got the brightest smile on her face because she just gave away her V-card to some dude because she ca n't read his mind ? As the main character of the story , I would 've hoped the show would do a little more to make her understandable and someone to invest your interest in , not someone you keep secretly hoping gets killed off or put into a coma . I ca n't find anything about her character that I like and even the fact that she can read minds is impressively \textbf{dreary} and not the least bit interesting . I will not be wasting my time with watching Season 2 come June .
\end{minipage} }
Correct label: negative. \\
Model confidence on original example: 79.4.
\end{figure*}

\begin{figure*}
Data augmentation model, example 2 \\
\fbox{ \begin{minipage}{\textwidth}
Original: Well this movie is \textbf{amazingly} \textbf{awful} . I felt sorry for the actors involved in this project because I 'm \textbf{sure} they did not write their lines . Which were sometimes delivered with slight sarcasm , which lead me to believe they were not taking this movie seriously , nor \textbf{could} anybody who watches this \textbf{obnoxious} off beat monster slasher . While watching this `` Creature Unknown '' I could not help but think that there was not much of a budget or a competent writer on the crew . But , if you go into watching this for a laugh you 'll be happy , the movie is \textbf{shameless} to mocking itself because i cant see how anybody \textbf{could} look at this and be proud of pumping this straight to DVD clichéd wan na be action thriller/horror movie fightfest to light .
\end{minipage} }
\fbox{ \begin{minipage}{\textwidth}
Perturbed: Well this movie is \textbf{marvellously} \textbf{horrifying} . I felt sorry for the actors involved in this project because I 'm \textbf{confident} they did not write their lines . Which were sometimes delivered with slight sarcasm , which lead me to believe they were not taking this movie seriously , nor \textbf{would} anybody who watches this \textbf{abhorrent} off beat monster slasher . While watching this `` Creature Unknown '' I could not help but think that there was not much of a budget or a competent writer on the crew . But , if you go into watching this for a laugh you 'll be happy , the movie is \textbf{cheeky} to mocking itself because i cant see how anybody \textbf{would} look at this and be proud of pumping this straight to DVD clichéd wan na be action thriller/horror movie fightfest to light .
\end{minipage} }
Correct label: negative. \\
Model confidence on original example: 97.8.
\end{figure*}

\begin{figure*}
Data augmentation model, example 3 \\
\fbox{ \begin{minipage}{\textwidth}
Original: Watched on Hulu ( far too many \textbf{commercials} ! ) so it \textbf{broke} the pacing but even still , it was like watching a really \textbf{bad} buddy movie from the early sixties . Dean Martin and Jerry Lewis where both parts are \textbf{played} by Jerry Lewis . If I were Indian , I 'd protest the portrayal of all males as venal and all women as shrews . They cheated for the music videos for western sales and \textbf{used} a lot of western models so the males could touch them I \textbf{usually} \textbf{enjoy} Indian films a lot but this was a major \textbf{disappointment} , especially for a modern Indian film . The \textbf{story} does n't take place in India ( the uncle keeps referring to when Mac will return to India ) but I ca n't find out where it is supposed to be happening .
\end{minipage} }
\fbox{ \begin{minipage}{\textwidth}
Perturbed: Watched on Hulu ( far too many \textbf{announcements} ! ) so it \textbf{cracked} the pacing but even still , it was like watching a really \textbf{wicked} buddy movie from the early sixties . Dean Martin and Jerry Lewis where both parts are \textbf{accomplished} by Jerry Lewis . If I were Indian , I 'd protest the portrayal of all males as venal and all women as shrews . They cheated for the music videos for western sales and \textbf{utilizes} a lot of western models so the males could touch them I \textbf{commonly} \textbf{savor} Indian films a lot but this was a major \textbf{frustration} , especially for a modern Indian film . The \textbf{history} does n't take place in India ( the uncle keeps referring to when Mac will return to India ) but I ca n't find out where it is supposed to be happening .
\end{minipage} }
Correct label: negative. \\
Model confidence on original example: 90.1.
\end{figure*}

\begin{figure*}
Data augmentation model, example 4 \\
\fbox{ \begin{minipage}{\textwidth}
Original: Weak start , solid middle , fantastic finish . That 's my impression of this film , anyway . I liked Simon Pegg in the two films I 've seen him in -- - Hot Fuzz , and Shaun of the Dead . His role here , though , took a completely different turn . Shows his range as an actor , but nonetheless I really disliked th character as he was portrayed at the beginning . There 's a kind of humour I call `` frustration \textbf{comedy} . '' Its supposed `` jokes '' and wit are really nothing more than painful and awkward moments . Much like the Bean character Rowan Atkinmson plays . There are a number of other comedic actors who portray similar characters too . I do n't mean to bash them here , so will not . But do be warned that if you are like me , and you dislike smarmy and maddeningly bungling idiots , Pegg shows just such characteristics for the first third of this film . It DOES get better , however . I read somewhere that this is based on a true story . Hmmm . Maybe . The film 's story stopped being annoying , and became kind of a triumph of the `` little guy '' in the final third . I do n't need all films to be sugar and light -- - but coincidentally , as this film got better , it also started to be more and more of a happy ending . It was also a pleasure to see an old favourite , Jeff Bridges , play a role so masterfully . I liked `` Iron Man , '' but was saddened by the fact that Bridges ' character was a villain . Purely personal taste , of course , as his acting in that was superb . Nonetheless , he was a marvel here as the Bigger Than Life man of vision , the publisher of Sharps . It was nice to see him in a role that I could actually enjoy . Overall then , I liked it ! I just wish I had come in 40 minutes late , and missed the beginning .
\end{minipage} }
\fbox{ \begin{minipage}{\textwidth}
Perturbed: Weak start , solid middle , fantastic finish . That 's my impression of this film , anyway . I liked Simon Pegg in the two films I 've seen him in -- - Hot Fuzz , and Shaun of the Dead . His role here , though , took a completely different turn . Shows his range as an actor , but nonetheless I really disliked th character as he was portrayed at the beginning . There 's a kind of humour I call `` frustration \textbf{charade} . '' Its supposed `` jokes '' and wit are really nothing more than painful and awkward moments . Much like the Bean character Rowan Atkinmson plays . There are a number of other comedic actors who portray similar characters too . I do n't mean to bash them here , so will not . But do be warned that if you are like me , and you dislike smarmy and maddeningly bungling idiots , Pegg shows just such characteristics for the first third of this film . It DOES get better , however . I read somewhere that this is based on a true story . Hmmm . Maybe . The film 's story stopped being annoying , and became kind of a triumph of the `` little guy '' in the final third . I do n't need all films to be sugar and light -- - but coincidentally , as this film got better , it also started to be more and more of a happy ending . It was also a pleasure to see an old favourite , Jeff Bridges , play a role so masterfully . I liked `` Iron Man , '' but was saddened by the fact that Bridges ' character was a villain . Purely personal taste , of course , as his acting in that was superb . Nonetheless , he was a marvel here as the Bigger Than Life man of vision , the publisher of Sharps . It was nice to see him in a role that I could actually enjoy . Overall then , I liked it ! I just wish I had come in 40 minutes late , and missed the beginning .
\end{minipage} }
Correct label: positive. \\
Model confidence on original example: 57.0.
\end{figure*}

\begin{figure*}
Data augmentation model, example 5 \\
\fbox{ \begin{minipage}{\textwidth}
Original: How The Grinch Stole Christmas instantly \textbf{stole} my heart and \textbf{became} my favorite \textbf{movie} almost from my very first \textbf{viewing} . Now , eight viewings later , it still \textbf{has} the same impact on me as it did the first time I \textbf{saw} it . Screenwriters Jeffery Price \& Peter S. Seaman of Who Framed Roger Rabbit \textbf{fame} do a \textbf{fantastic} job of \textbf{adapting} the story of The Grinch to the screen . Ron Howard 's direction brought the \textbf{story} to full \textbf{life} , and Jim Carrey 's typically \textbf{energetic} performance as The Grinch steals the show . Some detractors of the film have claimed that it is not true to the \textbf{spirit} or principles of the original story . Having read the original story , I \textbf{must} say I can not agree . The movie makes the very same point about Christmas and its true meaning as the original story . Indeed , it \textbf{enhances} the \textbf{impact} of the \textbf{story} by making it more personal by showing us how and why The Grinch \textbf{became} what he was . *MILD SPOILERS* ( They probably would n't ruin the movie for you ... but \textbf{if} you have n't \textbf{seen} it \textbf{yet} and you 're one of those who wants to know NOTHING about a story until you 've seen it , you should skip the next two paragraphs . ) I think just about everyone can relate to The Grinch 's \textbf{terrible} \textbf{experiences} in school . I think all of us , at one time or another , were the \textbf{unpopular} one in school who was always \textbf{picked} on . I know I was ... and that 's why I personally had so much sympathy for The Grinch and what he went through . And Cindy Lou Who 's naive idealism , believing that nobody can be all \textbf{bad} , was heart rending . When \textbf{everyone} else had turned their backs on The Grinch out of \textbf{fear} and \textbf{ignorance} , Cindy Lou was determined to be his friend . If only everyone could have such an attitude . In fact , I think the only thing that might 've \textbf{made} the film a little better would have been to \textbf{further} tone down the adult \textbf{humor} and content . It was already \textbf{pretty} \textbf{restrained} , but any of this \textbf{adult} \textbf{humor} ( like when The Grinch slammed nose first into Martha May Whovier 's cleavage ) just does n't fit in a story like this . This one 's well on its way to being a Christmas classic , \textbf{taking} a \textbf{richly} \textbf{deserved} place alongside the book and the Chuck Jones cartoon as a must-see of \textbf{every} Christmas season .
\end{minipage} }
\fbox{ \begin{minipage}{\textwidth}
Perturbed: How The Grinch Stole Christmas instantly \textbf{stolen} my heart and \textbf{went} my favorite \textbf{films} almost from my very first \textbf{opinion} . Now , eight viewings later , it still \textbf{had} the same impact on me as it did the first time I \textbf{watched} it . Screenwriters Jeffery Price \& Peter S. Seaman of Who Framed Roger Rabbit \textbf{celebrity} do a \textbf{wondrous} job of \textbf{adaptation} the story of The Grinch to the screen . Ron Howard 's direction brought the \textbf{storytelling} to full \textbf{lifetime} , and Jim Carrey 's typically \textbf{dynamic} performance as The Grinch steals the show . Some detractors of the film have claimed that it is not true to the \textbf{gist} or principles of the original story . Having read the original story , I \textbf{should} say I can not agree . The movie makes the very same point about Christmas and its true meaning as the original story . Indeed , it \textbf{improves} the \textbf{influence} of the \textbf{storytelling} by making it more personal by showing us how and why The Grinch \textbf{went} what he was . *MILD SPOILERS* ( They probably would n't ruin the movie for you ... but \textbf{whether} you have n't \textbf{noticed} it \textbf{though} and you 're one of those who wants to know NOTHING about a story until you 've seen it , you should skip the next two paragraphs . ) I think just about everyone can relate to The Grinch 's \textbf{awful} \textbf{experiment} in school . I think all of us , at one time or another , were the \textbf{unwanted} one in school who was always \textbf{pick} on . I know I was ... and that 's why I personally had so much sympathy for The Grinch and what he went through . And Cindy Lou Who 's naive idealism , believing that nobody can be all \textbf{horrid} , was heart rending . When \textbf{somebody} else had turned their backs on The Grinch out of \textbf{angst} and \textbf{ignorant} , Cindy Lou was determined to be his friend . If only everyone could have such an attitude . In fact , I think the only thing that might 've \textbf{brought} the film a little better would have been to \textbf{furthermore} tone down the adult \textbf{comedy} and content . It was already \textbf{abundantly} \textbf{scant} , but any of this \textbf{grownups} \textbf{comedy} ( like when The Grinch slammed nose first into Martha May Whovier 's cleavage ) just does n't fit in a story like this . This one 's well on its way to being a Christmas classic , \textbf{pick} a \textbf{meticulously} \textbf{deserving} place alongside the book and the Chuck Jones cartoon as a must-see of \textbf{any} Christmas season .
\end{minipage} }
Correct label: positive. \\
Model confidence on original example: 96.4.
\end{figure*}

\begin{figure*}
Data augmentation model, example 6 \\
\fbox{ \begin{minipage}{\textwidth}
Original: \textbf{so} ... it 's really sexist , and classist , and i thought that it might not be in the beginning stages of the movie , like when stella tells steven that she would really like to change herself and begin speaking in the right way and he tells her not to change . well , he certainly \textbf{changed} his tune , and it seems that the other reviewers followed suit . what at the beginning appears to be a love story is really about social placement and women as sacrificial mothers . the end of the movie does not make \textbf{her} a hero , it makes the whole thing \textbf{sad} . and its sad that people think it makes her a hero . perhaps that is the comment of the movie that people should take away . positive reception reflects continual patriarchal \textbf{currents} in the social conscience . \textbf{yuck} .
\end{minipage} }
\fbox{ \begin{minipage}{\textwidth}
Perturbed: \textbf{even} ... it 's really sexist , and classist , and i thought that it might not be in the beginning stages of the movie , like when stella tells steven that she would really like to change herself and begin speaking in the right way and he tells her not to change . well , he certainly \textbf{change} his tune , and it seems that the other reviewers followed suit . what at the beginning appears to be a love story is really about social placement and women as sacrificial mothers . the end of the movie does not make \textbf{his} a hero , it makes the whole thing \textbf{hapless} . and its sad that people think it makes her a hero . perhaps that is the comment of the movie that people should take away . positive reception reflects continual patriarchal \textbf{current} in the social conscience . \textbf{eww} .
\end{minipage} }
Correct label: negative. \\
Model confidence on original example: 83.0.
\end{figure*}

\begin{figure*}
Data augmentation model, example 7 \\
\fbox{ \begin{minipage}{\textwidth}
Original: Michael Kallio gives a strong and convincing performance as Eric Seaver , a troubled young \textbf{man} who was horribly mistreated as a little boy by his \textbf{monstrous} , abusive , alcoholic stepfather Barry ( a \textbf{genuinely} \textbf{frightening} \textbf{portrayal} by Gunnar Hansen ) . Eric has a compassionate fiancé ( sweetly played by the \textbf{lovely} Tracee Newberry ) and a \textbf{job} transcribing autopsy reports at a local morgue . Haunted by his bleak past , egged on by the bald , beaming Jack the demon ( a truly \textbf{creepy} Michael Robert Brandon ) , and sent over the edge by the recent death of his mother , Eric goes off the deep end and embarks on a brutal killing spree . Capably directed by Kallio ( who also wrote the tight , astute script ) , with uniformly fine acting by a sound no-name cast ( Jeff Steiger is especially good as Eric 's wannabe helpful guardian angel Michael ) , rather rough , but \textbf{overall} polished cinematography by George Lieber , \textbf{believable} true-to-life characters , jolting outbursts of raw , \textbf{shocking} and unflinchingly \textbf{ferocious} violence , a \textbf{moody} , \textbf{spooky} score by Dan Kolton , an uncompromisingly \textbf{downbeat} ending , grungy Detroit , Michigan \textbf{locations} , a grimly serious \textbf{tone} , and a taut , gripping narrative that stays on a steady track throughout , this extremely potent and gritty \textbf{psychological} horror thriller makes for often absorbing and disturbing viewing . A real sleeper .
\end{minipage} }
\fbox{ \begin{minipage}{\textwidth}
Perturbed: Michael Kallio gives a strong and convincing performance as Eric Seaver , a troubled young \textbf{guy} who was horribly mistreated as a little boy by his \textbf{atrocious} , abusive , alcoholic stepfather Barry ( a \textbf{honestly} \textbf{terrible} \textbf{description} by Gunnar Hansen ) . Eric has a compassionate fiancé ( sweetly played by the \textbf{handsome} Tracee Newberry ) and a \textbf{work} transcribing autopsy reports at a local morgue . Haunted by his bleak past , egged on by the bald , beaming Jack the demon ( a truly \textbf{horrible} Michael Robert Brandon ) , and sent over the edge by the recent death of his mother , Eric goes off the deep end and embarks on a brutal killing spree . Capably directed by Kallio ( who also wrote the tight , astute script ) , with uniformly fine acting by a sound no-name cast ( Jeff Steiger is especially good as Eric 's wannabe helpful guardian angel Michael ) , rather rough , but \textbf{general} polished cinematography by George Lieber , \textbf{plausible} true-to-life characters , jolting outbursts of raw , \textbf{appalling} and unflinchingly \textbf{brutish} violence , a \textbf{lunatic} , \textbf{terrible} score by Dan Kolton , an uncompromisingly \textbf{dismal} ending , grungy Detroit , Michigan \textbf{placements} , a grimly serious \textbf{tones} , and a taut , gripping narrative that stays on a steady track throughout , this extremely potent and gritty \textbf{psychiatric} horror thriller makes for often absorbing and disturbing viewing . A real sleeper .
\end{minipage} }
Correct label: positive. \\
Model confidence on original example: 99.9.
\end{figure*}

\begin{figure*}
Data augmentation model, example 8 \\
\fbox{ \begin{minipage}{\textwidth}
\small
Original: When a \textbf{stiff} turns up with pneumonic plague ( a variant of bubonic plague ) , U.S. Public Health Service official Dr. Clinton Reed ( Richard Widmark ) immediately quarantines everyone whom he knows was near the body . Unfortunately , the stiff got that way by being murdered , and there 's a good chance that the murderer will start spreading the plague , leading to an epidemic . Enter Police Captain Tom Warren ( Paul Douglas ) , who is \textbf{enlisted} to track down the murderer as soon as possible and avert a \textbf{possible} national disaster . While Panic in the Streets is a quality film , it suffers from being slightly unfocused and a bit too sprawling ( my \textbf{reason} for bringing the score down to an eight ) . It wanders the genres from noirish gangster to medical disaster , \textbf{police} procedural , thriller and even romance . This is not director Elia Kazan 's \textbf{best} work , but \textbf{saying} that is a bit disingenuous . Kazan is the helmer \textbf{responsible} such masterpieces as A Streetcar Named Desire ( 1951 ) , On The Waterfront ( 1954 ) and East of Eden ( 1955 ) , after all . This film predates those , but Kazan \textbf{has} said that he was already `` untethered '' by the studio . Taking that \textbf{freedom} too far may partially account for the sprawl . The film is set in New Orleans , a city where Kazan `` used to wander around . . . night and day so I knew it well '' . He wanted to exploit the environment . `` It 's so terrific and \textbf{colorful} . I wanted boats , steam \textbf{engines} , \textbf{warehouses} , jazz joints -- all of New Orleans '' . Kazan handles each genre of Panic in the Streets well , but they could be connected better . The film would have benefited by staying with just one or two of its \textbf{moods} . The sprawl in terms of setting would have \textbf{still} worked . Part of the \textbf{dilemma} may have been caused by the fact that Panic in the Streets was an attempt to merge two stories by writers Edna and Edward Anhalt , `` Quarantine '' and `` Some Like 'Em Cold '' . The gangster material , which ends up in firmly in thriller territory with an extended chase scene near the end of the \textbf{film} , is probably the highlight . Not surprisingly , Kazan has said that he believes the villains are `` more colorful -- I never had much affection for the good guys anyway . I do n't like puritans '' . A close second is the only material that approaches the `` panic '' of the title -- the discovery of the plague and the \textbf{attempts} to track down the \textbf{exposed} , inoculate them and contain the \textbf{disease} . While there is plenty of suspense during these two `` \textbf{moods} '' , much of the film is also a \textbf{fairly} straightforward drama , with pacing more \textbf{typical} of that genre . The dialogue throughout is excellent . The stylistic \textbf{difference} to many modern films could hardly be more pronounced . It is intelligent , delivered \textbf{quickly} and well enunciated by each character . Conflict is n't created by `` \textbf{dumb} '' decisions but smart \textbf{moves} ; events and characters ' actions are more like a chess game . When unusual stances are taken , such as Reed withholding the plague from the \textbf{newspapers} , he gives \textbf{relatively} lengthy justifications for his decisions , which other characters argue over . In light of this , it 's \textbf{interesting} that Kazan believed that `` \textbf{propriety} , religion , \textbf{ethics} and the middle class are all murdering us '' . That idea works its way into the film through the alterations to the norm , or allowances away from it , made by the protagonists . For example , head gangster Blackie ( Jack Palance in his first film role ) is offered a `` Get Out of Jail Free '' card if he 'll cooperate with combating the plague . The technical \textbf{aspects} of the film are \textbf{fine} , if nothing exceptional , but the real reasons to watch are the performances , the intriguing \textbf{scenario} and the well-written dialogue .
\end{minipage} }
\fbox{ \begin{minipage}{\textwidth}
\small
Perturbed: When a \textbf{rigid} turns up with pneumonic plague ( a variant of bubonic plague ) , U.S. Public Health Service official Dr. Clinton Reed ( Richard Widmark ) immediately quarantines everyone whom he knows was near the body . Unfortunately , the stiff got that way by being murdered , and there 's a good chance that the murderer will start spreading the plague , leading to an epidemic . Enter Police Captain Tom Warren ( Paul Douglas ) , who is \textbf{recruited} to track down the murderer as soon as possible and avert a \textbf{probable} national disaster . While Panic in the Streets is a quality film , it suffers from being slightly unfocused and a bit too sprawling ( my \textbf{justification} for bringing the score down to an eight ) . It wanders the genres from noirish gangster to medical disaster , \textbf{cops} procedural , thriller and even romance . This is not director Elia Kazan 's \textbf{better} work , but \textbf{arguing} that is a bit disingenuous . Kazan is the helmer \textbf{liable} such masterpieces as A Streetcar Named Desire ( 1951 ) , On The Waterfront ( 1954 ) and East of Eden ( 1955 ) , after all . This film predates those , but Kazan \textbf{have} said that he was already `` untethered '' by the studio . Taking that \textbf{freely} too far may partially account for the sprawl . The film is set in New Orleans , a city where Kazan `` used to wander around . . . night and day so I knew it well '' . He wanted to exploit the environment . `` It 's so terrific and \textbf{colored} . I wanted boats , steam \textbf{motors} , \textbf{stores} , jazz joints -- all of New Orleans '' . Kazan handles each genre of Panic in the Streets well , but they could be connected better . The film would have benefited by staying with just one or two of its \textbf{passions} . The sprawl in terms of setting would have \textbf{however} worked . Part of the \textbf{stalemate} may have been caused by the fact that Panic in the Streets was an attempt to merge two stories by writers Edna and Edward Anhalt , `` Quarantine '' and `` Some Like 'Em Cold '' . The gangster material , which ends up in firmly in thriller territory with an extended chase scene near the end of the \textbf{movie} , is probably the highlight . Not surprisingly , Kazan has said that he believes the villains are `` more colorful -- I never had much affection for the good guys anyway . I do n't like puritans '' . A close second is the only material that approaches the `` panic '' of the title -- the discovery of the plague and the \textbf{attempt} to track down the \textbf{unmasked} , inoculate them and contain the \textbf{ailment} . While there is plenty of suspense during these two `` \textbf{passions} '' , much of the film is also a \textbf{reasonably} straightforward drama , with pacing more \textbf{symptomatic} of that genre . The dialogue throughout is excellent . The stylistic \textbf{variance} to many modern films could hardly be more pronounced . It is intelligent , delivered \textbf{fast} and well enunciated by each character . Conflict is n't created by `` \textbf{idiotic} '' decisions but smart \textbf{move} ; events and characters ' actions are more like a chess game . When unusual stances are taken , such as Reed withholding the plague from the \textbf{journal} , he gives \textbf{comparatively} lengthy justifications for his decisions , which other characters argue over . In light of this , it 's \textbf{fascinating} that Kazan believed that `` \textbf{validity} , religion , \textbf{ethos} and the middle class are all murdering us '' . That idea works its way into the film through the alterations to the norm , or allowances away from it , made by the protagonists . For example , head gangster Blackie ( Jack Palance in his first film role ) is offered a `` Get Out of Jail Free '' card if he 'll cooperate with combating the plague . The technical \textbf{matters} of the film are \textbf{handsome} , if nothing exceptional , but the real reasons to watch are the performances , the intriguing \textbf{screenplay} and the well-written dialogue .
\end{minipage} }
Correct label: positive. \\
Model confidence on original example: 89.9.
\end{figure*}

\begin{figure*}
Data augmentation model, example 9 \\
\fbox{ \begin{minipage}{\textwidth}
Original: i \textbf{completely} agree with jamrom4.. this was the single \textbf{most} \textbf{horrible} movie i have ever seen.. \textbf{holy} \textbf{crap} it was terrible.. i was warned not to see it..and \textbf{foolishly} i \textbf{watched} it anyway.. about 10 minutes \textbf{into} the \textbf{painful} experience i \textbf{completely} gave up on watching the atrocity..but sat through until the end..just to \textbf{see} \textbf{if} i could.. \textbf{well} i \textbf{did} and now i wish i had not..it was disgusting..nothing happened and the ending was \textbf{all} preachy..no \textbf{movie} that \textbf{bad} has the right to survive..i implore all of you to spare \textbf{yourself} the terror of \textbf{fatty} \textbf{drives} the bus..if only i \textbf{had} heeded the \textbf{same} warning..please save \textbf{yourself} from this movie..i have a feeling those who rated it \textbf{highly} were \textbf{involved} in the making of the movie..and \textbf{should} \textbf{all} be wiped off the face of the planet..
\end{minipage} }
\fbox{ \begin{minipage}{\textwidth}
Perturbed: i \textbf{fully} agree with jamrom4.. this was the single \textbf{greatest} \textbf{horrifying} movie i have ever seen.. \textbf{saintly} \textbf{shit} it was terrible.. i was warned not to see it..and \textbf{recklessly} i \textbf{saw} it anyway.. about 10 minutes \textbf{in} the \textbf{arduous} experience i \textbf{fully} gave up on watching the atrocity..but sat through until the end..just to \textbf{behold} \textbf{whether} i could.. \textbf{bah} i \textbf{got} and now i wish i had not..it was disgusting..nothing happened and the ending was \textbf{everybody} preachy..no \textbf{cinematic} that \textbf{wicked} has the right to survive..i implore all of you to spare \textbf{yourselves} the terror of \textbf{fat} \textbf{driving} the bus..if only i \textbf{has} heeded the \textbf{similar} warning..please save \textbf{himself} from this movie..i have a feeling those who rated it \textbf{supremely} were \textbf{embroiled} in the making of the movie..and \textbf{must} \textbf{everyone} be wiped off the face of the planet..
\end{minipage} }
Correct label: negative. \\
Model confidence on original example: 99.4.
\end{figure*}

\begin{figure*}
Data augmentation model, example 10 \\
\fbox{ \begin{minipage}{\textwidth}
Original: jim carrey can do anything . i thought this was going to be some \textbf{dumb} \textbf{childish} movie , and it TOTALLY was not . it was so incredibly funny for EVERYONE , adults \& kids . i \textbf{saw} it once cause it was almost out of theatres , and now it 's FINALLY coming out on DVD this tuesday and i 'm way to excited , as you can see . you should definitely see it if you have n't already , it was so \textbf{great} ! Liz
\end{minipage} }
\fbox{ \begin{minipage}{\textwidth}
Perturbed: jim carrey can do anything . i thought this was going to be some \textbf{idiotic} \textbf{puerile} movie , and it TOTALLY was not . it was so incredibly funny for EVERYONE , adults \& kids . i \textbf{noticed} it once cause it was almost out of theatres , and now it 's FINALLY coming out on DVD this tuesday and i 'm way to excited , as you can see . you should definitely see it if you have n't already , it was so \textbf{awesome} ! Liz
\end{minipage} }
Correct label: positive. \\
Model confidence on original example: 83.3.
\end{figure*}

\end{document}